\documentclass[10pt,journal,compsoc,cspaper]{IEEEtran}

\usepackage[dvips]{color}
\usepackage{mdframed}
\usepackage{dashrule}
\usepackage[cmex10]{amsmath}

\ifCLASSOPTIONcompsoc
\usepackage[tight,normalsize,sf,SF]{subfigure}
\else
\usepackage[tight,footnotesize]{subfigure}
\fi

\usepackage{graphicx}
\usepackage{epsfig}

\begin{document}

\title{Parallel D2-Clustering: Large-Scale Clustering of Discrete Distributions}

\author{Yu Zhang, 
        James Z. Wang,~\IEEEmembership{Senior Member, IEEE,} 
	and Jia Li,~\IEEEmembership{Senior Member, IEEE}%
\IEEEcompsocitemizethanks{\IEEEcompsocthanksitem Y. Zhang and J. Z. Wang are with the 
College of Information Sciences and Technology, The Pennsylvania State University,
University Park, PA 16802. J. Z. Wang was also with the National Science Foundation
when the work was done.
Email: \{yzz123, jwang\}@ist.psu.edu.
\IEEEcompsocthanksitem J. Li is with the Department of Statistics, The Pennsylvania State University,
University Park, PA 16802. She is also with the National Science Foundation. 
Email: jiali@stat.psu.edu.\protect\\
Any opinion, findings, and conclusions or recommendations expressed in this material are those of the authors and do not necessarily reflect the views of the National Science Foundation.}
\thanks{}}

\markboth{}%
{Shell \MakeLowercase{\textit{et al.}}: Bare Demo of IEEEtran.cls for Computer Society Journals}

\IEEEcompsoctitleabstractindextext{%
\begin{abstract}
The discrete distribution clustering algorithm, namely D2-clustering, has
demonstrated its usefulness in image classification and annotation
where each object is represented by a bag of weighed vectors.  The high
computational complexity of the algorithm, however, limits its
applications to large-scale problems.  We present a parallel
D2-clustering algorithm with substantially improved scalability.  A
hierarchical structure for parallel computing is devised
to achieve a balance between the individual-node computation
and the integration process of the algorithm.  Additionally, it is
shown that even with a single CPU, the hierarchical structure
results in significant speed-up. Experiments on real-world large-scale
image data, Youtube video data, and protein sequence data demonstrate the
efficiency and wide applicability of the parallel D2-clustering algorithm.  The loss in
clustering accuracy is minor in comparison with the original sequential algorithm.
\end{abstract}

}

\maketitle

\IEEEdisplaynotcompsoctitleabstractindextext

\IEEEpeerreviewmaketitle

\section{Introduction} \label{sec:intro}
\IEEEPARstart{T}{he} transportation problem is about finding an optimal matching between
elements in two sets with different Radon measures under a certain cost function.
The problem was initially formulated
by French mathematician G. Monge in the late 18th century~\cite{monge1781memoire},
and later
by Russian mathematician L. Kantorovich in 1942~\cite{kantorovich1942transfer}.
The solution of a transportation problem defines a metric between two distributions,
which is normally known as  the Kantorovich-Wasserstein metric.
With the wide applications of the transportation problem in different fields, there are
several variants, as well as names, of the corresponding transportation metric.
Readers can refer to~\cite{rachev1998mass} for these applications.

The Kantorovich-Wasserstein metric is better known as the Mallows
distance~\cite{mallows1972note} in the statistics community because
C.~L. Mallows used this metric (with $L_2$-norm as the cost function)
to prove some asymptotic property.
In the computer science community, the metric is often
called the Earth Mover's Distance (EMD)~\cite{rubner1998metric}.
Hereafter we refer to
the Kantorovich-Wasserstein metric as the Mallows distance.  It is not
until the recent decade that the Mallows distance has received much
attention in data mining and machine learning.

The advantages of the Mallows distance over a usual histogram-based distance
include its effectiveness with sparse representations and its consideration of 
cross-term relationships. We have found abundant applications of the
Mallows distance in video classification and retrieval~\cite{takada2008web,xu2008video}, 
sequence categorization~\cite{kosakovsky2010evolutionary}, 
and document retrieval~\cite{wan2007novel}, where objects are represented
by discrete distributions. The Mallows distance is
also adopted in a state-of-the-art real-time image annotation system named
ALIPR~\cite{li2007real}, in which a clustering algorithm on discrete
distributions was developed.

Clustering is a major unsupervised learning methodology for data
mining and machine learning. The K-means algorithm is one of the most
widely used algorithms for clustering.  To a large extent, its appeal
comes from its simple yet generally accepted objective function for
optimization.  The original K-means proposed by MacQueen~\cite{macqueen1967some}
only applies to the Euclidean space, which is a fundamental
limitation. Though there are quite a few attempts to extend K-means to
various distances other than the Euclidean distance~\cite{kashima2008k, mao1996self,merugu2004clustering},
these K-means' variants
generally work on vector data.
In many emerging applications,
however, the data to be processed are not vectors. 
In particular, bags of weighted vectors, which 
can be formulated as discrete distributions
with finite but arbitrary support, are frequently used as
descriptors for objects in information and image retrieval~\cite{datta2008image}. 
The discrete distribution clustering
algorithm, also known as the {\em D2-clustering}~\cite{li2007real},
tackles this type of data by optimizing an objective function defined
in the same spirit as that used by K-means, specifically, to minimize
the sum of distances between each object and its nearest
centroid.

Because linear programming is needed in D2-clustering and the number of
unknown variables grows with the number of objects in a cluster, the
computational cost would normally increase polynomially with the
sample size (assuming the number of clusters is roughly fixed).  
For instance, ALIPR takes several minutes to learn each semantic category 
by performing D2-clustering on $80$ images in it. However, it will be difficult 
for ALIPR to include over $1,000$ training images in one semantic category, 
because on a CPU with a clock speed of 2 GHz, the category may 
demand more than a day to cluster. There are some alternative
ways~\cite{julien2008image,yavlinsky2005automated} 
to perform the clustering faster, but they
generally perform some indirect or sub-optimal clustering instead of
the optimization based D2-clustering.

The clustering problems researchers tackle today are often of very
large scale. For instance, we are seeking to perform learning and
retrieval on images from the Web containing millions of pictures
annotated by one common word. This leads to demanding clustering
tasks even if we perform learning only on a fraction of the Web
data. Adopting bags of weighted vectors as descriptors, 
we will encounter the same issue in other applications such as video, biological sequence,
and text clustering, where the amount of data can be enormous.
In such cases, D2-clustering is computationally impractical.

Given the application potential of D2-clustering and its theoretical
advantage, it is of great interest to develop a variant of the
algorithm with a substantially lower computational complexity,
especially under a parallel computing environment.  With the
increasing popularity of large-scale cloud computing in information systems, we
are well motivated to exploit parallel computing.

\subsection{Preliminaries}\label{sec:pre}

D2-clustering attempts to minimize the
total squared Mallows distance between the data and the centroids.  Although
this optimization criterion is also used by K-means, the computation
in D2-clustering is much more sophisticated due to the nature of the
distance.  The D2-clustering algorithm is similar to K-means in that
it iterates the update of the partition and the update of the
centroids.  There are two stages within each iteration.  First, the
cluster label of each sample is assigned to be the label of its
closest centroid; and secondly, centroids are updated by minimizing
the within-cluster sum of squared distances.  To update the partition,
we need to compute the Mallows distance between a sample and each
current centroid.  To update the centroids, an embedded iterative
procedure is needed which involves a large-scale linear
programming problem in every iteration.  The number of unknowns
in the optimization grows with the number of samples in the
cluster.

Suppose $v_1,v_2,\ldots,v_{n'}$ are discrete distributions within a
certain cluster currently and
$v_i=\{(v_i^{(1)},p_{v_i}^{(1)}),\ldots,(v_i^{(t_i)},p_{v_i}^{(t_i)})\}$,
$i=1,2,\ldots,n'$,  
the following optimization is pursued
in order to update the centroid
$z=\{(z^{(1)},p_z^{(1)}),\ldots,(z^{(s')},p_z^{(s')})\}$.
\vspace{-0.05in}
\begin{equation*}
	\begin{aligned}
	&\varepsilon=\min_{z^{(r)},p_z^{(r)}:1 \leq r\leq s'}\sum_{i=1}^{n'}D^2(z,v_i) 
\end{aligned}
\vspace{-0.05in}
\end{equation*}
where
\vspace{-0.05in}
\begin{equation*}
	\begin{aligned}
& D^2(z,v_i)=
{\min_{w_{r,\alpha}^{(i)}}{\sum_{r=1}^{s'}{\sum_{\alpha=1}^{t_i}{w_{r,\alpha}^{(i)}\|z^{(r)}-v_i^{(\alpha)}\|^2}}}}\;,\\
\end{aligned}
\vspace{-0.05in}
\end{equation*}

subject to:
\vspace{-0.05in}
\begin{equation}
	\begin{aligned}
	&p_z^{(r)}\geq 0, r=1,\ldots,s'\;,\;
	\sum_{r=1}^{s'}{p_z^{(r)}}=1\;,\\
	&\sum_{\alpha=1}^{t_i}{w_{r,\alpha}^{(i)}}=p_z^{(r)}, i=1,\ldots,n', r=1,\ldots,s'\;,\\
	&\sum_{r=1}^{s'}{w_{r,\alpha}^{(i)}}=p_{v_i}^{(\alpha)}, i=1,\ldots,n', \alpha=1,\ldots,t_i\;,\\
	&w_{r,\alpha}^{(i)}\geq 0, i=1,\ldots,n', \alpha=1,\ldots,t_i, r=1,\ldots,s'\;.
\end{aligned}
	\label{D2opt}
\vspace{-0.05in}
\end{equation}

If $z$ is fixed, $D(z,v_i)$ is in fact the Mallows distance between $z$ and $v_i$.
However in (\ref{D2opt}), we cannot separately optimize $D(z,v_i)$, $i=1$, ..., $n'$, 
because $z$ is involved and
the variables $w^{(i)}_{r,\alpha}$ for different $i$'s affect each other
through $z$.  To solve the complex optimization problem,
the D2-clustering algorithm adopts the following
iterative strategy.

\vspace{\baselineskip}
{\noindent \bf \centering \small Algorithm 1. Centroid Update Process of D2-Clustering.\\}
\noindent\rule[0.5\baselineskip]{\columnwidth}{1.5pt}
{\small
\begin{list}{\labelitemi}{\leftmargin=0.5em}
\item[\bfseries{Step 1}] Fix $z^{(r)}$'s and update $p_z^{(r)}$, $r=1$, ..., $s'$,
and $w_{r,\alpha}^{(i)}$, $i=1$, ..., $n'$, $r=1$, ..., $s'$, $\alpha=1$, ...,
$t_i$. In this case the optimization~(\ref{D2opt}) of $p_z^{(r)}$'s and $w_{r,\alpha}^{(i)}$'s
reduces to a linear programming.
\item[\bfseries{Step 2}] Then fix $p_z^{(r)}$'s and $w_{r,\alpha}^{(i)}$'s and update $z^{(r)}, r=1$, ..., $s'$.
In this case the solution of $z^{(r)}$'s to optimize (\ref{D2opt}) are simply weighted averages:
\vspace{-0.05in}
\begin{equation}
	\begin{aligned}
z^{(r)}=\frac{\sum_{i=1}^{n'}{\sum_{\alpha=1}^{t_{i}}{w_{r,\alpha}^{(i)}v_i^{(\alpha)}}}}{\sum_{i=1}^{n'}{\sum_{\alpha=1}^{t_{i}}{w_{r,\alpha}^{(i)}}}}, r=1,\ldots, s'.
\end{aligned}
	\label{UpdateZ}
\vspace{-0.05in}
\end{equation}
\item[\bfseries{Step 3}] Calculate the objective function $\varepsilon$ in (\ref{D2opt}) using the
updated $z^{(r)}$'s, $p_z^{(r)}$'s, and $w_{r,\alpha}^{(i)}$'s. Terminate the iteration
until $\varepsilon$ converges. Otherwise go back to Step 1.
\end{list}
}
\noindent\rule[0.5\baselineskip]{\columnwidth}{1.5pt}

The linear programming in Step 1 is the primary cost in the centroid update process.
It solves $p_z^{(r)}$, $w_{r,\alpha}^{(i)}$,
$r=1$, ..., $s'$, $i=1$, ..., $n'$, $\alpha=1$, ..., $t_i$, a total of
$s'+s'\sum_{i=1}^{n'}{t_i}\simeq n's'^2$ parameters when $s'$ is set to be
the average of the $t_i$'s.  The number of constraints is
$1+n's'+\sum_{i=1}^{n'}{t_i}\simeq 2n's'$.  Hence, the number of
parameters in the linear programming grows linearly with the size of
the cluster.  Because it costs polynomial time to solve the linear
programming problem~\cite{karmarkar1984new}, D2-clustering is much
more complex than K-means algorithm in computation.

The analysis of the time complexity of K-means remains an unsolved
problem because the number of iterations for convergence is difficult
to determine.  Theoretically, the worst case time complexity for
K-means is ${\Omega(2^{\sqrt{n}})}$~\cite{arthur2006slow}. Arthur et
al.~\cite{arthur2009k} show recently that K-means has polynomial
smoothed complexity, which reflects the fact that K-means will
converge in a relatively short time in real cases (although no tight
upper or lower bound of time complexity has been proved). In practice,
K-means usually converges fast, and seldom costs an exponential number
of iterations. In many cases K-means converges in linear or even
sub-linear time~\cite{duda2001pattern}. Although there is still a gap
between the theoretical explanation and practice, we can consider K-means
an algorithm with at most polynomial time.

Although the D2-clustering algorithm interlaces the update of clustering
labels and centroids in the same manner as K-means, the update of centroids
is much more complicated.  As described above, to update each
centroid involves an iterative procedure where a large-scale linear
programming problem detailed in (\ref{D2opt}) has to be solved in
every round.  This makes D2-clustering usually polynomially more
complex than K-means.

The computational intensiveness of D2-clustering limits its usages to only
small-scale problems. With emerging demands to extend the algorithm to
large-scale datasets, e.g., online image datasets, video resources, and
biological databases, we exploit parallel processing in a cluster computing environment
in order to overcome the inadequate scalability of D2-clustering.


\subsection{Overview of Parallel D2-Clustering}

To reduce the computational complexity of D2-clustering, we seek a
new parallel algorithm.  Parallelization is a widely adopted strategy
for accelerating algorithms.  There are several recent
algorithms~\cite{chang2009parallel,song2008parallel,wang2009plda} that
parallelize large-scale data processing. By distributing parallel
processes to different computation units, the overall processing time
can be greatly reduced.

The labeling step of D2-clustering can be easily parallelized because the 
cluster label of each point can be optimized separately. The centroid update 
stage described by (\ref{D2opt}), which is the performance bottleneck
of D2-clustering, makes the parallelization difficult. 
Unlike K-means' centroid update, which
is essentially a step of computing an average, 
the computation of a centroid in D2-clustering is 
far more complicated than any linear combination of the data points. Instead,
all the data points in the corresponding cluster will play a role in the
linear programming problem in (\ref{D2opt}).

The averaging step of K-means can be naturally parallelized. We divide the data
into segments, and compute a local average of each segment in parallel.
All the local averages can then be combined to a global average 
of the whole group of data. This is usually how K-means is being 
parallelized~\cite{dhillon2000data,kantabutra2000parallel}.
To the best of our knowledge, there is no mathematical 
equivalence of a parallel algorithm for the linear programming problem
 (\ref{D2opt}) in D2-clustering. However, we can adopt a similar strategy to
parallelize the centroid update in D2-clustering: dividing the data into segments 
based on their adjacency, computing some local centroids for each segment in parallel, 
and combining the local centroids to a global centroid.
Heuristically we can get a good estimation of the optimal centroid
in this way, considering that the notion of centroid is exactly
for a compact representation of a group of data.  Hence, a centroid
computed from the local centroids should represent well the entire
cluster, although some loss of accuracy is expected. 

The most complex part in the parallelization of D2-clustering is the combination
of local centroids. Same as the computation of local centroids, 
the combination operation also involves a linear programming
with polynomial time in terms of data size. To achieve a low overall runtime, 
we want to keep both linear programming problems at a small scale, which cannot be
achieved simultaneously when the size of the overall data is large. As a result,
we propose a new hierarchical structure to conduct parallel computing
recursively through multiple levels.  This way, we can control the
scale of linear programming in every segment and at every level.

\begin{figure}[t!]
	\begin{center}
		\epsfig{file=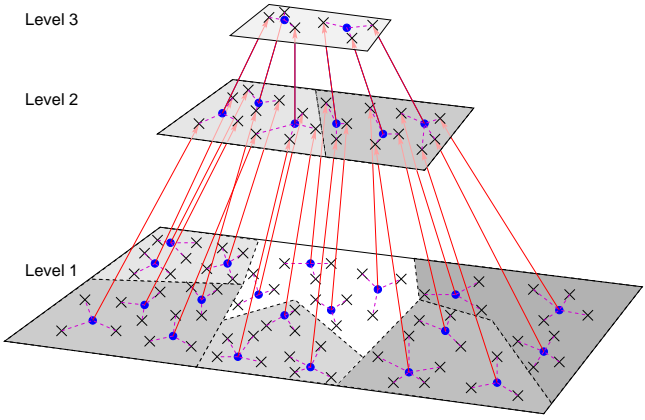, width=0.47\textwidth}
	\end{center}
	\caption{An illustration of the structure of the parallel
D2-clustering algorithm. Crossings are data to be clustered at each
level. Different shadings denote different segments to perform local
D2-clustering, which can be done in parallel within a level. Dots
are cluster centroids. Centroids at a certain level become data points
at the next level. The clustering stops going to a higher level until
we get the desired number of clusters.  }
	\label{fig:parallelD2}
	\vspace{-0.1in}
\end{figure}

We demonstrate the structure of the algorithm in Fig.~\ref{fig:parallelD2},
which will be explained in detail in Section~\ref{sec:alg_str}.
In short, we adopt a hierarchically structured parallel algorithm.
The large dataset is first clustered into segments, within each of
which D2-clustering is performed in parallel.  The centroids generated
for each segment are passed to the next level of the hierarchy and are
treated as data themselves.  The data size in the new level can still
be large, although it should be much smaller than the original data
size.  At the new level, the same process is repeated to cluster data
into segments and perform D2-clustering in parallel for each segment.
When the amount of data at a certain level is sufficiently small, no
further division of data is needed and D2-clustering on all the data
will produce the final result.  We can think of the granularity of the
data becoming coarser and coarser when the algorithm traverses through
the levels of the hierarchy. Details about each step will be introduced in
the following sections.

Particularly for implementation, parallel computing techniques, such
as Message Passing Interface (MPI)~\cite{thakur2005optimization} and
MapReduce~\cite{dean2008mapreduce} can be applied. In the current
work, MPI is used to implement the parallel algorithm.

\subsection{Outline of the Paper}
The remainder of paper is organized as follows.  The parallel
D2-clustering algorithm and its performance analysis are presented in
Section~\ref{sec:alg}.  Experimental results and comparison with other
methods are described in Section~\ref{sec:exp}.  Finally, we conclude
in Section
\ref{sec:conclude}.

\section{Parallel D2-clustering Algorithm} \label{sec:alg}

In this section, we present the parallel D2-clustering algorithm.
Hereafter, we refer to the original D2-clustering as the {\it
  sequential algorithm}.  We will describe the hierarchical structure
for speeding up D2-clustering and a weighted version of D2-clustering
which is needed in the parallel algorithm but not the sequential.  We
will also describe in detail the coordination among the individual 
computation nodes (commonly called the slave
processors) and the node that is responsible for integrating the 
results (called the master processor).  Finally, we analyze the
computational complexity of the parallel algorithm and explain various
choices made in the algorithm in order to achieve fast speed.

\subsection{Hierarchical Structure} \label{sec:alg_str}

Fig.~\ref{fig:parallelD2} illustrates the hierarchical structure used
for parallel processing.  
All input data are denoted by crossings at the bottom level (Level 1)
in Fig.~\ref{fig:parallelD2}.  Because of the large data size, these
data points (each being a discrete distribution) are divided into
several segments (reflected by different shadings in the figure) which
are small enough to be clustered by the sequential D2-clustering.
D2-clustering is performed within each segment.  Data points in each
segment are assigned with different cluster labels, and several
cluster centroids (denoted by blue dots in the figure) are obtained.

These cluster centroids are regarded as good summarization for the
original data.  Each centroid approximates the effect of the data
assigned to the corresponding cluster, and is taken to the next level
in the hierarchy.  Obviously, the number of data points to be
clustered at the second level is much reduced from the original data
size although it can still be large for sequential D2-clustering.
Consequently, the data may be further divided into small segments
which sequential D2-clustering can handle.  In our implementation, we
keep the size of a segment at different levels roughly the same.  When
the algorithm proceeds to higher levels, the number of segments
decreases.  The hierarchy terminates when the data at a level can be
put in one segment.

In short, at each level, except the bottom level, we perform
D2-clustering on the centroids acquired from the previous level.
Fig.~\ref{fig:parallelD2} shows that the data size at the third level
is small enough for not proceeding to another level.  Borrowing the
terminology of a tree graph, we can call a centroid a parent point and
the data points assigned to its corresponding cluster child points.
The cluster labels of the centroids at a higher level are inherited by
their child points, and propagate further to all the descendent points.
To help visualize the tree, in Fig.~\ref{fig:parallelD2}, data points
in one cluster at a certain level are connected to the corresponding
centroid using dashed lines. And the red arrows between levels
indicate the correspondence between a centroid and a data point at the
next level.  To determine the final clustering result for the original
data, we only need to trace each data point through the tree and find
the cluster label of its ancestor at the top level.

In such a hierarchical structure, the number of centroids becomes
smaller and smaller as the level increases. Therefore it is guaranteed
that the hierarchy can terminate.  Also since D2-clustering is
conducted within small segments, the overall clustering can be
completed fast.  More detailed discussion about the complexity of the
algorithm will be provided in a later section.

\subsection{Initial Data Segmentation}\label{sec:DataSegment}

We now describe the initial data segmentation method. In this step, we want to partition the dataset
into groups all smaller than a certain size.
The partitioning process
is in a similar spirit as the initialization step of the
LBG algorithm proposed by Linde \textit{et al.}~\cite{linde1980algorithm}, an early instance of K-means
in signal processing.
The whole dataset is iteratively split into partitions
until a stopping criterion is satisfied.
And within each iteration, the splitting
is an approximation of, and computationally less expensive than, an optimal clustering.

\begin{figure}[t!]
	\begin{center}
		\epsfig{file=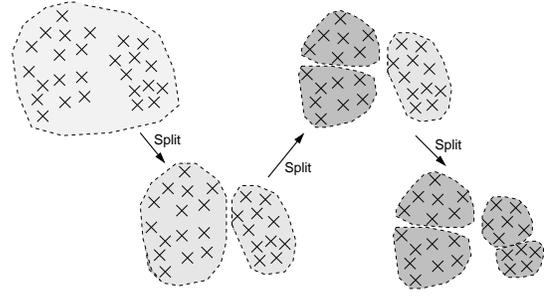, width=0.8\columnwidth}
	\end{center}
	\caption{Initial data segmentation. Each time the largest segment is split
	into two clusters, until all segments are small enough.}
	\label{fig:init}
	\vspace{-0.1in}
\end{figure}

Fig. \ref{fig:init} illustrates the iterative process containing a series of binary splitting.
Different from the initialization
of the LBG algorithm, which splits every segment into two
parts, doubling the total number of segments
in every iteration, our approach only splits one segment
in each round.
The original data are split into two groups using a binary clustering approach (to be introduced later). 
Then we further split the segments by the
same process repetitively. Each time we split the segment containing 
the most data points into two clusters, so on so forth. The splitting process will stop
when all the segments are smaller than a certain size.

As shown by Fig. 2, in each round of dividing a group of data
into two segments, we aim at good clustering so that
data points in each segment are close.
On the other hand, this clustering step
needs to be fast since it is a preparation before conducting
D2-clustering in parallel.  The method we use is a computationally
reduced version of D2-clustering. It is suitable for clustering
bags of weighted vectors. Additionally, it is fast because of some
constraints imposed on the centroids.  We refer to this version as the
{\it constrained D2-clustering}.

The constrained D2-clustering is a reduced version of D2-clustering.
We denote each data point to be clustered as
$v_i=\{(v_i^{(1)},p_{v_i}^{(1)}),\ldots,(v_i^{(t_i)},p_{v_i}^{(t_i)})\},
i=1,\ldots,n$, and $k$ initial centroids as
$z_j=\{(z_j^{(1)},p_{z_j}^{(1)}),\ldots,(z_j^{(s_j)},p_{z_j}^{(s_j)})\},
j=1,\ldots,k$. Parameters $t_i$ and $s_j$ are the numbers of support vectors
in the corresponding discrete distributions $v_i$ and $z_j$. And
$C(i)\in\{1,\ldots,k\}$ denotes the cluster label assigned to $v_i$. 
The D2-clustering and constrained D2-clustering algorithms are described below.

\vspace{\baselineskip}
{\noindent \bf \centering \small Algorithm 2. D2-Clustering and Constrained D2-Clustering.\\}
\noindent\rule[0.5\baselineskip]{\columnwidth}{1.5pt}
{\small
\begin{list}{\labelitemi}{\leftmargin=0.5em}
        \item[\bfseries{Step 1 }] Select some random centroids as the initialization.
	\item[\bfseries{Step 2 }]
Allocate a cluster label to each data point $v_i$ by selecting the
		nearest centroid.\\ $C(i)=\arg\min_{j}{D(v_i,z_j)}$,\\
		where $D(v_i,z_j)$ is the Mallows distance between $v_i$ and $z_j$.

	\item[\bfseries{Step 3 }] {\bfseries(For D2-clustering only)}
Solve (\ref{D2opt}) to update $z_j$, the centroid of each cluster, using the data points
$\{v_{i:C(i)=j}\}$ assigned to the cluster. Algorithm 1, which is also an iterative process,
is invoked here to solve $z_j$.

\item[\bfseries{Step 3*}] {\bfseries(For constrained D2-clustering only)}
        Keep $p_{z_j}^{(r)}=1/s_j, r=1,\ldots,s_j$ fixed when invoking Algorithm 1 to solve $z_j$. The linear programming
       (\ref{D2opt}) can thus be separated into several smaller ones and runs much faster.
        The result of $z_j$ will be a uniform weighted bag of vectors containing support vectors at optimized locations. 
	\item[\bfseries{Step 4 }]
                Suppose the objective function in (\ref{D2opt}) converges to $\varepsilon_j$ after updating $z_j$.
               Calculate the overall objective function for the clustering:\\
		$\varepsilon(C)=\sum_{j=1}^{k}{\varepsilon_j}$.\\ If $\varepsilon(C)$ converges, the
		clustering finishes;\\otherwise go back to Step 2.

\end{list}
}
\noindent\rule[0.5\baselineskip]{\columnwidth}{1.5pt}

The difference between D2-clustering and constrained D2-clustering is the
way to update centroids in Step~3 of the above algorithm. As discussed previously
in Section \ref{sec:pre}, we need to solve a large-scale linear programming problem
in Algorithm 1 to update each $z_j$ in D2-clustering.
The linear programming involves the optimization of  $p_{z_j}^{(r)}$ and $w_{r,\alpha}^{(i)}$,
$r=1,\ldots,s_j$, $\alpha=1,\ldots,t_i$, $i: C(i)=j$
($w_{r,\alpha}^{(i)}$ is the matching weight between the $r$-th vector of $z_j$ and the $\alpha$-th vector of $v_i$).
This makes D2-clustering computationally complex.

In the constrained D2-clustering algorithm, 
we replace Step 3 by Step 3* in the D2-clustering algorithm.
We simplify the optimization of D2-clustering by
assuming uniform $p_{z_j}^{(r)}=1/s_j$, $r=1,\ldots, s_j$, for any
$j=1,\ldots,k$.  With this simplification, in the linear programming
problem in Step 1 of Algorithm 1, we only need to solve $w_{r,\alpha}^{(i)}$'s. 
Moreover, $w_{r,\alpha}^{(i)}$'s can be optimized separately for different
$i$'s, which significantly reduces the number of parameters in a
single linear programming problem.  More specifically, instead of
solving one linear programming problem with a large number of
parameters, we solve many linear programming problems with a small
number of parameters.  Due to the usual polynomial complexity of the
linear programming problems, this can be done much faster than the original
D2-clustering.

\subsection{Weighted D2-clustering}
Data segments generated by the initialization step are distributed to
different processors in the parallel algorithm. Within each processor,
a D2-clustering is performed to cluster the segment of data.
Such a process is done at different levels as illustrated in Fig. \ref{fig:parallelD2}.

Clustering at a certain level usually produces unbalanced clusters
containing possibly quite different numbers of data points.  If we
intend to keep equal contribution from each original data point, the
cluster centroids passed to a higher level in the hierarchy should be
weighted and the clustering method should take those weights into
account.  We thus extended D2-clustering to a weighted version.  As
will be shown next, the extension is straightforward and results in
little extra computation.

It is obvious that with weighted data, the step of nearest neighbor assignment
of centroids to data points will not be affected, while the update of
the centroids under a given partition of data will.  Let us inspect
the optimization in (\ref{D2opt}) for updating the centroid $z$ of one
cluster.  When data $v_i$ have weights $\omega_i$, a natural
extension of the objective function is
\vspace{-0.05in}
\begin{equation}
	\min_{z^{(r)},p_z^{(r)}:1 \leq r\leq s'}\sum_{i=1}^{n'}\omega_i D^2(z,v_i) 
	\label{WeightedD2}
\vspace{-0.05in}
\end{equation}
where $D(z,v_i)$ is defined in (\ref{D2opt}).

In Step 1 of Algorithm 1, we fix $z_j^{(r)}$, $r=1,\ldots,s_j$, and solve a linear
programming problem with objective function (\ref{WeightedD2}) under the
same constraints specified in (\ref{D2opt}) to get the vector weights $p_{z_j}^{(r)}$'s
and a matching $\omega_{r,\alpha}^{(i)}$ between $z_j$ and each $v_i$, $i:C(i)=j$. Then in Step 2, the update for
$z_j^{(r)}$ is revised to a weighted version:
\vspace{-0.05in}
\begin{equation}
z_j^{(r)}=\frac{\sum_{i:C(i)=j}{\omega_i\sum_{\alpha=1}^{t_{i}}{w_{r,\alpha}^{(i)}v_i^{(\alpha)}}}}{\sum_{i:C(i)=j}{\omega_i\sum_{\alpha=1}^{t_{i}}{w_{r,\alpha}^{(i)}}}}\;.
\label{WeightedUpdate}
\vspace{-0.05in}
\end{equation}

Because the optimal solution of centroid is not affected when the
weights are scaled by a common factor, we simply use the number of
original data points assigned to each centroid as its weight.  The
weights can be computed recursively when the algorithm goes through
levels in the hierarchy.  At the bottom level, each original data point is
assigned with weight $1$.  The weight of a data point in a parent
level is the sum of weights of its child data points.

\subsection{Algorithmic Description}
With details in several major aspects already explained in the
previous sections, we now present the overall flow of the parallel
D2-clustering algorithm. 
\begin{figure}[ht!]
	\begin{center}
		\epsfig{file=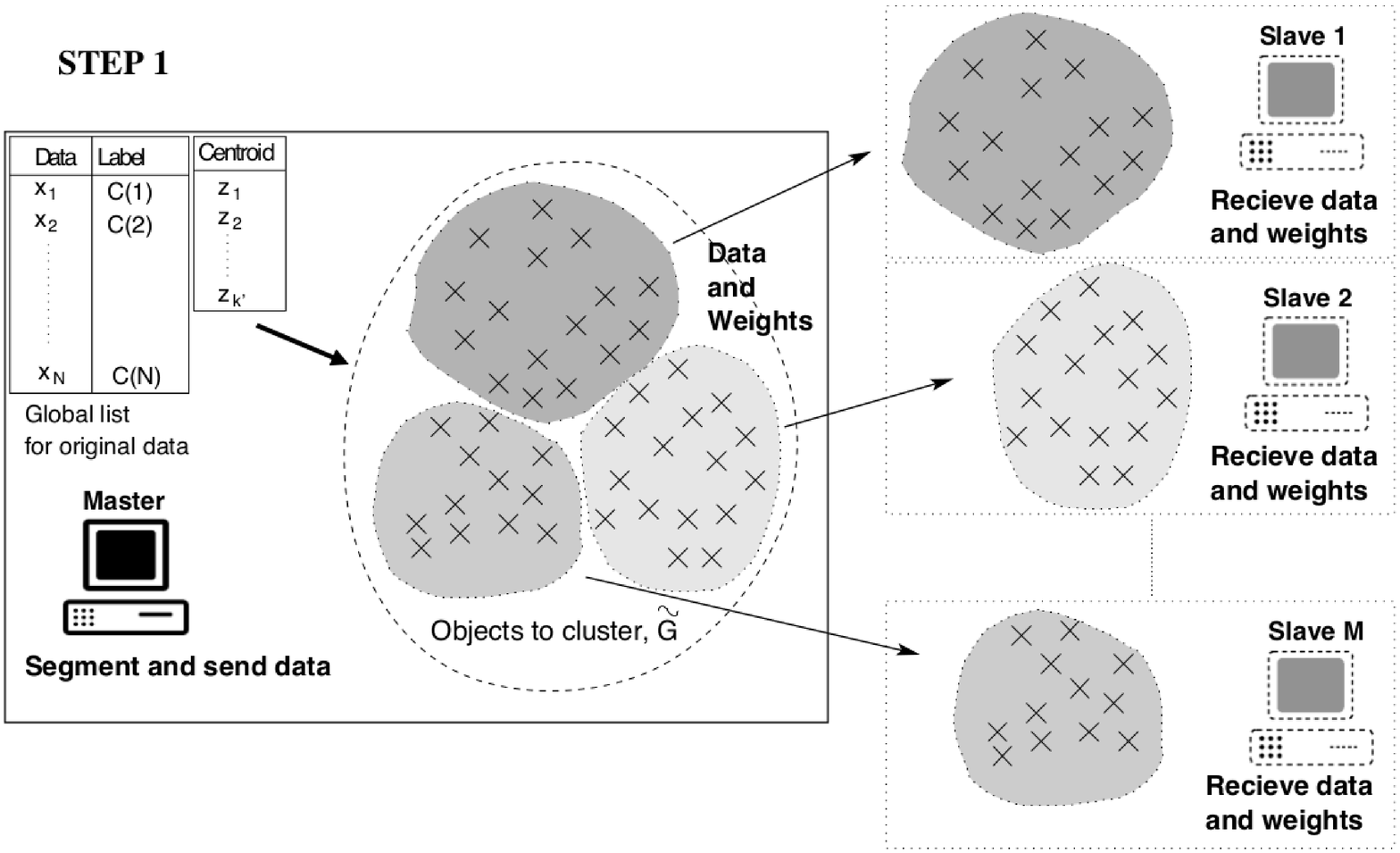, width=3.2in}\\
\noindent\hdashrule[0.5\baselineskip]{0.95\columnwidth}{1pt}{3pt}
		\epsfig{file=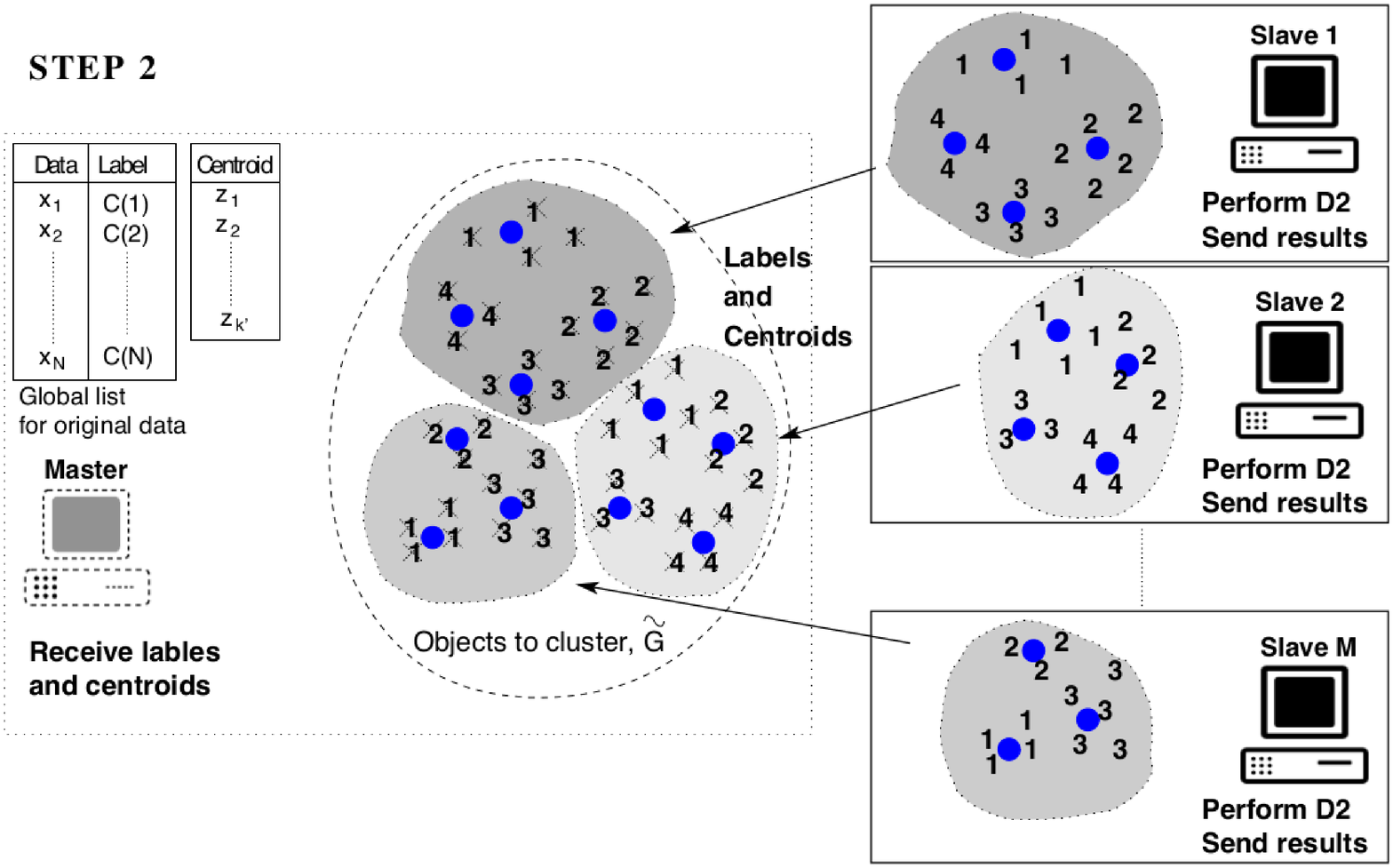, width=3.2in}\\
\noindent\hdashrule[0.5\baselineskip]{0.95\columnwidth}{1pt}{3pt}
		\epsfig{file=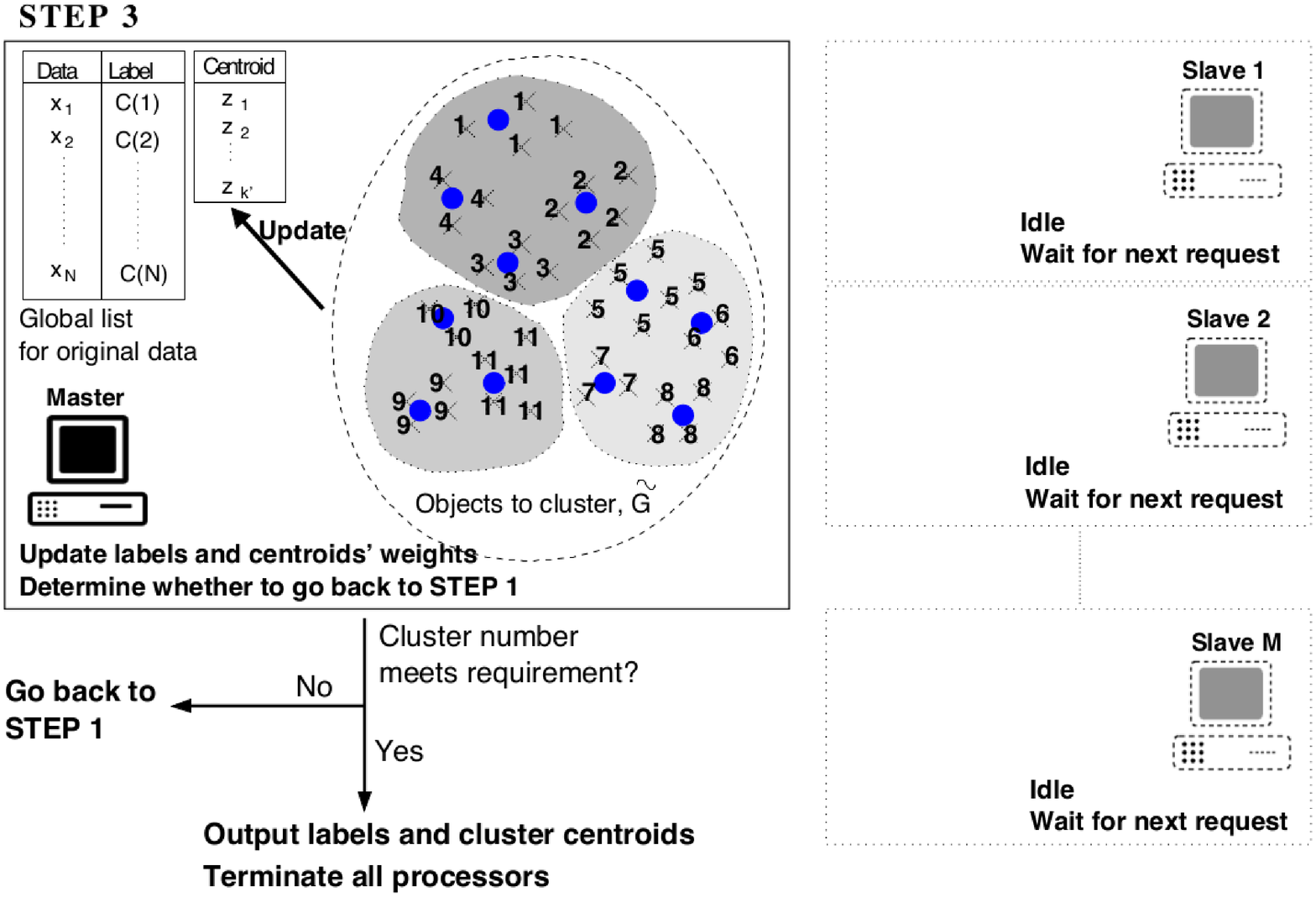, width=3.25in}\\
	\end{center}
\vspace{-0.1in}
	\caption{Work flow and communication among master and slave processors in parallel D2-clustering
        for one iteration. 
	Within each iteration (level), there are three synchronized steps
        for the master and slave processors. In each step, one of the master or slave side is performing
        computation (marked in solid line) and the other is idle and waiting for incoming data and message
        (marked in dashed line).} 
	\label{fig:parallelcompute}
	\vspace{-0.1in}
\end{figure}
The parallel program runs on a computation cluster with multiple
processors. In general, there is a ``master'' processor that controls
the flow of the program and some other ``slave'' processors in charge
of computation for the sub-problems. In the implementation of parallel D2-clustering,
the master processor will perform the initial data segmentation at each level
in the hierarchical structure specified in Fig.~\ref{fig:parallelD2},
and then distribute the data segments to different slave processors.
Each slave processor will cluster the data it received by the weighted D2-clustering
algorithm independently, and send the clustering result back to the master processor.
As soon as the master processor receives all the within-segment clustering
results from the slave processors, it will combine these results and
set up the data for clustering at the next level if necessary.

The current work adopts MPI for implementing the parallel program.
In order to keep a correct work flow for both master and slave
processors, we use synchronized communication ways in MPI to perform
message passing and data transmission.
A processor can {\it broadcast} its local data to all others, and {\it send} data
to (or {\it receive} data from) a specific processor. 
Being synchronized, all these operations will not return until 
the data transmission is completed. Further, we can {\it block} the
processors until all have reached a certain blocking point. In this way,
different processors are enforced to be synchronized.

Because the communication is synchronized, we can divide the overall
work flow into several steps for both the master and slave processors.
To better explain the algorithm, in particular how the master 
and slave processors interact, we describe the working processes of 
both the master and slaves together based on these steps, 
although they are essentially different sets of instructions
performed at different CPU cores respectively.

The work flow is illustrated in Fig.~\ref{fig:parallelcompute}.
There are one master processor (on the left of the figure) and 
$M$ slave processors (on the right of the figure). The master processor
will read the original set of discrete distributions
$X=\{x_1,x_2,\ldots,x_N\}$ and maintain the clustering label
$C(i) \in \{1,...,k'\}$ for each data entry $x_{i}$, as well as the
corresponding cluster centroids $Z=\{z_1, z_2,..., z_{k'}\}$.
Initially the weight of each data entry is set to be $\omega_i=1$, $i=1,\dots,N$.
At each level, there are three steps. In each step, either the master processor
or the slave processors are idle, passively waiting and receiving incoming data,
and the other one is performing computation (which are marked by dashed
and solid icons in the figure respectively). Suppose the targeted number of clusters
is $k$ and the sequential (weighted) D2-clustering is applied to segments
of data containing no more than $\tau$ entries. The algorithm has a hierarchical structure
and iteratively performs clustering level by level. And each iteration contains
the following three steps.

\vspace{\baselineskip}
{\noindent \bf\centering\small Algorithm 3. Parallel D2-Clustering.\\}
\noindent\rule[0.5\baselineskip]{\columnwidth}{1.5pt}
{\small
\begin{list}{\labelitemi}{\leftmargin=0.5em}
	\item[\bfseries{Step 1}] {\bf Data segmentation in the master's side.} 
\begin{list}{$\bullet$}{\leftmargin=1em}
     \item  {\bf \emph{The master processor}} segments the data objects $\tilde{G}$
                (containing $n$ objects) at current level
                into $m$ groups, $G_1$, ..., $G_{m}$,
                each of which contains no more than $\tau$ entries, 
                using the constrained D2-clustering introduced in Section \ref{sec:DataSegment}.
                The indices of data entries assigned to segment
                $G_{\mu}$ are stored in array $L_{\mu}$. 
                For each segment $G_{\mu}$,  the master processor sends the segment index $\mu$, data in
	        the segment and their weights to the slave processor with index $(\mu \mod M)$.

        \item        {\bf \emph{Each slave processor}} will listen to and receive any incoming data segment 
		from the master processor in this step. It will store 
                data in the segment
                $V_{\mu}=\{v_{\mu,1},v_{\mu,2},\ldots,v_{\mu,n_{\mu}}\}$ and the corresponding weights
	        $W_{\mu}=\{\omega_{\mu,1},\omega_{\mu,2},\ldots,\omega_{\mu,n_{\mu}}\}$ locally for 
                clustering. In addition, the index of the corresponding segment, $\mu$, is also transmitted and kept in
                the slave processor. Since the number of data segments 
                $m$ might be larger than the number of slave
                processors $M$, the master processor uses a modular operation to distribute different segments
                to different slave processors and each slave processor may handle multiple segments sequentially.
                The segment indices help the master processor manage the data distribution and identify
                the result of a certain segment when it is returned by a slave processor in the next step.
\end{list}
	\item[\bfseries{Step 2}] {\bf Parallel D2-clustering in the slaves' side.}
\begin{list}{$\bullet$}{\leftmargin=1em}
              \item {\bf \emph{The master processor}} listens to the clustering results, including both the labels
                 $C_{\mu}$ (containing $n_{\mu}$ labels) and centroids $Z_{\mu}$ (containing
	$k_{\mu}$ centroids), as well as the index $\mu$, for each segment, and is blocked until it receives all segments' results.
             \item {\bf \emph{Each slave processor}} will perform weighted D2-clustering on $V_{\mu}$ with      
              weights $W_{\mu}$
	      by Algorithm 1 (using (\ref{WeightedD2}) and (\ref{WeightedUpdate})) when
	      updating centroids. Then a set of centroids
	      $Z_{\mu}=\{z_{\mu,1},\ldots,z_{\mu,k_{\mu}}\}$ and cluster labels 
	     $C_{\mu}=\{C_{\mu}(1),\ldots,C_{\mu}(n_{\mu})\}$ are acquired, 
	     where $C_{\mu}(i)=j$ if $v_{\mu,i}$
	     is assigned to centroid $z_{\mu,j}$.
             After this it sends back the segment index $\mu$, 
             the set of centroids $Z_{\mu}$, and
             cluster labels $C_{\mu}$ to the master processor.
\end{list}
	\item[\bfseries{Step3}] {\bf Work flow control in the master's side.}
\begin{list}{$\bullet$}{\leftmargin=1em}
     \item           {\bf \emph{The master processor}} will merge the clustering results of all the segments
                        and then determine whether to proceed to the next level.\\
                        The current overall centroid list is set to $Z=\emptyset$ at the beginning.
                        The centroids $Z_{\mu}$ for group $G_{\mu}$ are appended to $Z$ one by one. And 
                        the labeling list of $\tilde{G}$, $\tilde{C}(i), i=1,...,n$,
                        is updated once a new group of centroids is being added.
                        Let the most recent list $Z$ before adding $Z_{\mu}$ be $Z^{(\mu-1)}$
                       ($Z^{(0)}=\emptyset$). $\tilde{C}(\cdot)$ is computed by (\ref{UpdateLabel}).
\vspace{-0.05in}
                       \begin{eqnarray}
                       \tilde{C}(L_{\mu}(i))=C_{\mu}(i)+|Z^{(\mu-1)}|, i=1,...,n_{\mu} \;.
                       \label{UpdateLabel}
\vspace{-0.05in}
                        \end{eqnarray}
                       After all $Z_{\mu}$ is merged to $Z$ and all $\tilde{C}(\cdot)$ is computed,
                        we update the cluster label of a data entry in the
                       original dataset (bottom level), ${C}(i)$, 
                       by inheriting label from its ancestor in $\tilde{G}$.
\vspace{-0.05in}
                       \begin{eqnarray}
                       C(i)=\tilde{C}(C(i)), i=1,...,N \;.
                       \label{UpdateGlobalLabel}
\vspace{-0.05in}
                        \end{eqnarray}
                       Assume $k_{\mu}$ is the number of clusters in segment $G_{\mu}$ and
                       $k'=\sum_{\mu=1}^{m}k_{\mu}$.
                       At the end of the current level of clustering,
                       we get a list of centroids, $Z=\{z_1, z_2,..., z_{k'}\}$,
                       and the cluster labels $C(1), C(2), ..., C(N)$ for the original data.
                      
                        If $k' \leq k$, clustering is accomplished.  
                        The master processor will output $Z$ and $C$, and sends a 
                        termination signal to all the processors.

                        Otherwise clustering at a higher level is needed. Let 
                        $\omega'_i=\omega_i$, $i=1,\ldots,n$.
                        Update $\tilde{G}=Z$ and assign new weights to each entry in
                        $\tilde{G}$ by
                        $\omega_i=\sum_{\alpha:\tilde{C}(\alpha)=i}\omega'_\alpha$,
                        $i=1,\ldots,k'$.
			Then the master processor will go back to Step 1 and send signals to slave processors
			to tell them also to return to Step 1. 
     \item          {\bf \emph{The slave processors}} are blocked in this step, doing nothing until the 
               master processor tells them the next action, either to return to Step 1 and receive new data, 
		or to terminate.
\end{list}
\end{list}
}
\rule[0.5\baselineskip]{\columnwidth}{1.5pt}
\subsection{Run-Time Complexity}\label{sec:SpeedGain}

In this section, we analyze the time complexity of the parallel
D2-clustering algorithm.

As described in Section~\ref{sec:intro}, for D2-clustering, the
 computational complexity of updating the centroid for one cluster in
 one iteration is polynomial in the size of the cluster.  The total
 amount of computation depends on computation within one iteration as
 well as the number of iterations.  Again, as discussed in
 Section~\ref{sec:intro}, even for K-means clustering, the
 relationship between the number of iterations needed for convergence
 and the data size is complex.  In practice, we often terminate the
 iteration after a pre-specified number of rounds (e.g., 500) if the
 algorithm has not converged before that.  As a result, in the
 analysis of computational complexity, we focus on the time needed to
 complete one iteration and assume the time for the whole algorithm is
 a multiple of that, where the multiplication factor is a large constant.

Suppose there are $N$ data entries in total and we want to cluster
them into $k$ clusters.  Assuming the clusters are of similar sizes,
each cluster contains roughly $\frac{N}{k}$ data entries.
The sequential D2-clustering algorithm will run in approximately
$O\left(k(\frac{N}{k})^{\gamma}\right)$ time, where $\gamma$ is the degree of
the polynomial time required by linear programming for computing the
centroid of one cluster. In many applications, when $N$ increases, the
number of clusters $k$ may stay the same or increase little.  As a
result, the time for D2-clustering increases polynomially with $N$.

Now consider the time required by the parallel D2-clustering.  The
sequential D2-clustering is then performed on segments of data with
sizes no larger than $\tau$.  Each segment is clustered into $c$
groups with roughly equal size $e$.  Thus, $\tau=ce$.  After
clustering every data segment, the number of data to be processed at
the next level of the hierarchy shrinks from $N$ to
$\frac{N}{e}$, since we use each cluster centroid to
summarize data entries assigned to the corresponding cluster.

Based on the run-time analysis on the sequential clustering, the time
for clustering one segment of data is $O(ce^{\gamma})=O(\tau e^{\gamma-1})$. The
 combined CPU time used for clustering all segments is then $O(N
 e^{\gamma-1})$. The actual time for performing clustering also depends on
 the number of available parallel CPU cores. Assuming there are $M+1$
 CPUs, where $M$ is the number of CPUs performing slave clustering operations
 and there is one CPU serving as the master.  When $M\geq \frac{N}{\tau}$,
 it can be guaranteed that each segment is distributed to a unique
 CPU.  Therefore the actual time used is $O(\tau e^{\gamma-1})$.  When
 $M<\frac{N}{\tau}$, there will be some CPUs handling more than one
 clustering jobs sequentially, so the actual time is $O\left(\frac{N}{M}
 e^{\gamma-1}\right)$.

In addition to clustering the segments, it also takes time to divide
data into segments by the approach described in Section
\ref{sec:DataSegment}. The data segments are generated by a binary
tree-structured clustering process.  For the simplicity of analysis,
we assume the tree is balanced.  The time to split a node is linear to
the number of data entries in the node; and the time for
creating $\frac{N}{\tau}$ segments is $O\left(N
\log\frac{N}{\tau}\right)$.  Similarly, there is also some computation for
initialization when clustering each segment, which takes
$O\left(\tau\log\frac{\tau}{e}\right)$ time. When $\tau$ is large, the
initialization cannot be neglected for run-time analysis.  In summary,
the overall actual time for performing clustering at the first level in
Fig.~\ref{fig:parallelD2} is
\vspace{-0.05in}
\begin{equation}
	T(1)=O\left(N \log\frac{N}{\tau}+\frac{e^{\gamma-1}+\log\frac{\tau}{e}}{\min{(M,\frac{N}{\tau})}} N\right)\;.
	\label{equ:timecomp}
\vspace{-0.05in}
\end{equation}

The hierarchical clustering may not reach the desired number of
clusters at the first level, where $\frac{N}{e}$ cluster centroids are acquired.
If $\frac{N}{e}>k$, the algorithm proceeds to
the next level, where each cluster centroid will be treated as a data entry.
If we keep the same values for parameters $\tau$, $e$, and $M$,
the time for the second level clustering can be
 calculated simply using (\ref{equ:timecomp}) with $N$ replaced by
$\frac{N}{e}$.  More generally, we can obtain the time complexity for any
$l$-th level, $l\geq 2$, by
\vspace{-0.05in}
\begin{equation}
	\hspace{-0.02in}T(l)=O\left(\frac{N}{e^{l-1}} \log\frac{N}{\tau e^{l-1}}+\frac{e^{\gamma-1}+\log\frac{\tau}{e}}{\min{(M,\frac{N}{\tau e^{l-1}})}e^{l-1}} N\right)\;.
	\label{equ:timelevel}
\vspace{-0.05in}
\end{equation}

Obviously $T(l)>eT(l+1)$. The algorithm will terminate at level $L$
when $\frac{N}{e^{L-1}}\leq k$.  Hence, the highest level $L=\lceil
\log_t \frac{N}{k} \rceil$. And the total run-time is
$T(1)+\ldots+T(L)<\frac{e}{e-1} T(1)\leq 2 T(1)$.  The last inequality
comes from the fact $e\geq 2$ (If $e=1$, there is essentially no
clustering).  We thus see that the computational order of the total
run-time is the same as that of $T(1)$.

With fixed $N$ and $\tau$, the complexity $T(1)$ in
(\ref{equ:timecomp}) indicates that a small value of $e$ is
favored.  Another benefit of having a small $e$ is that a centroid can
better represent a small number of data entries.  We set $e=5$ in our
implementation.  We do not go as low as $e=2$ because more levels will
be needed which makes the coordination among processors harder to track.

To achieve small $T(1)$, (\ref{equ:timecomp}) shows that there is
a trade-off for $\tau$.  Given the number of slave CPUs, $M$, which is
inherently determined by the hardware, we try to choose a sufficiently
large $\tau$ so that the number of segments $\frac{N}{\tau}$ is no
greater than $M$.  This ensures that no CPU needs to process multiple
segments sequentially.  Moreover, large $\tau$ will reduce the time
for generating the segments, which corresponds to the first term in
(\ref{equ:timecomp}).  On the other hand, when $\tau$ is large,
assuming $\frac{N}{\tau}\leq M$, the second term in
(\ref{equ:timecomp}) is in the linearithmic order of $\tau$.  Hence, $\tau$
should not be unnecessarily large.  If $M$ slave processors are
guaranteed, $\tau=\frac{N}{M}$ is a good choice.  In our
implementation, we observe that the sequential D2-clustering can
handle 50 data entries and 10 clusters (5 entries per cluster) at
satisfactory speed.  We thus set $\tau=50$ and $e=5$ in the parallel
algorithm by default.

Comparing the two terms in (\ref{equ:timecomp}), the first term
dominates when $N$ is large. Otherwise, the second
term dominates because $e^{\gamma-1}$ can be substantial.  Assuming
parameters $M$, $\tau$, and $e$ are fixed, when $\frac{N}{\tau}\leq
M$, $T(1)$ will be dominated by a constant given in the second term.
As $N$ grows, $T(1)$ will eventually be dominated by the first term,
which is a linearithmic order ($O(N\log N)$).

Based on the analysis above, we can see the speedup of the parallel
D2-clustering over the original sequential algorithm, which runs in
time $O\left(\frac{N^{\gamma}}{k^{\gamma-1}}\right)$. Usually $N\gg k$, so the
sequential algorithm has to solve a large linear program
when $N$ is large. In the parallel algorithm, during the update of a
centroid, on average $e$ data entries are handled, rather than
$\frac{N}{k}$.  There is thus no large-scale linear programming.  Even
when there is only one CPU, i.e., $M=1$, (that is, essentially no
parallel processing), the ``parallel'' algorithm based on the
hierarchy will have run-time $O\left(N\left(e^{\gamma-1}+\log\frac{N}{e}\right)\right)$, which is
linearithmic order in $N$ rather than polynomial order.  If we have
enough CPUs to guarantee $M$ is always no smaller than
$\frac{N}{\tau}$, the second term of (\ref{equ:timecomp}) is always a
constant with respect to $N$.  When $N$ is large, the first term
$O\left(N\log\frac{N}{\tau}\right)$ dominates, which is much lower in order than
$O\left(\frac{N^{\gamma}}{k^{\gamma-1}}\right)$.


\section{Experiments}\label{sec:exp}

To validate the advantage and usefulness of this algorithm, we
conduct experiments on four different types of dataset, including image,
video, protein sequence, and synthetic datasets. Though our work
was originally motivated by image concept learning, we apply
the parallel D2-clustering algorithm to different scenarios. Readers
in these related areas may therefore find applications that are
suitable to embed the algorithm into. 

First, we test the algorithm on images crawled from Flickr in
several experiments. When crawling images from
Flickr, we use the Flickr API to perform keyword query for certain
concepts, and download top $N$ images with highest values of
interestingness. The number of images $N$ is different in different
experiments. Typically it is at a scale of thousands, much larger than
the scale the sequential D2-clustering has been applied. 
 If such clustering is performed for multiple concepts, the
training dataset for image tagging can easily reach several millions,
which is quite large for image annotation.

For such image datasets, each image is first segmented into two sets of
regions based on the color (LUV components) and texture features
(Daubechies-4 wavelet~\cite{daubechies1992ten} coefficients)
respectively.  
We then use two bags of weighted vectors $U$ and $V$
, one for color and the other for texture, to describe an image as $I=(U,V)$.  
The distance between two images, $I_i=(U_i,V_i)$ and $I_j=(U_j,V_j)$,
is defined as
\vspace{-0.05in}
\begin{equation}
	\hat{D}(I_i,I_j)=(D^2(U_i,U_j)+D^2(V_i,V_j))^{\frac{1}{2}}\;,
	\label{equ:imgdist}
\vspace{-0.05in}
\end{equation}
where $D(\cdot,\cdot)$ is the Mallows distance. 
It is straightforward to extend the
D2-clustering algorithm to the case of multiple discrete distributions 
using the combined distance defined in (\ref{equ:imgdist}). See~\cite{li2007real}
for details. 
The order of time complexity increases simply by
a multiplicative factor equal to the number of distribution types, the
so-called super-dimension.

Second, we adopt videos queried and downloaded from Youtube.
We represent each video by a bag of weighted vectors, and conduct
parallel D2-clustering on these videos. Then we check the accordance between
the clustering result and the videos'~genre.

For the video clustering, we adopt the features used in~\cite{yanagawa2007columbia}.
Edge Direction Histogram (EDH), Gabor (GBR), and Grid Color Moment (GCM) features
are extracted from each frame. A video is segmented into several sub-clips
based on the continuity of these three features among nearby frames~\cite{lienhart1999comparison}.
Using the time percentage of each sub-clip as the weight of its average feature
vector, we represent the video by a bag of weighted vectors by combining all sub-clips.

Third, we download the SwissProt protein data~\cite{boeckmann2003swiss} and apply clustering
on a subset of this dataset. Prosite protein family data~\cite{hulo2006prosite} provides
the class labels of these protein sequences. 
Using the Prosite data, we can select protein sequences from several
certain classes as the experiment dataset,
which contains data gathering around some patterns.
Each protein sequence is composed of 20 types of amino acids.
We count the frequency of each amino acid in a sequence, and use the frequency
as the signature of the sequence.

At last, we synthesize some bags of weighted vectors following certain distribution
and cluster them.

The parallel program is
deployed on a computation cluster at The Pennsylvania State University named ``CyberStar''
consisting of  512 quad-core CPUs and
therefore 2048 computation units. Since $M=2048$, theoretically it can be
guaranteed that at each level every data segment can be processed in
parallel by a unique CPU core when the data size is several thousands.
In practice the system will put a limit to the maximal number of
processors each user can occupy at a time because it is a publicly shared server.

In this section, we first evaluate objectively the D2-clustering algorithm on image datasets with
different topics and scales. Then we show the D2-clustering results for the image, video, protein,
and synthetic datasets~respectively.

\begin{figure*}[t]
\begin{center}
	\subfigure[]{
		\label{fig:sos}
		\includegraphics[width=0.31\textwidth]{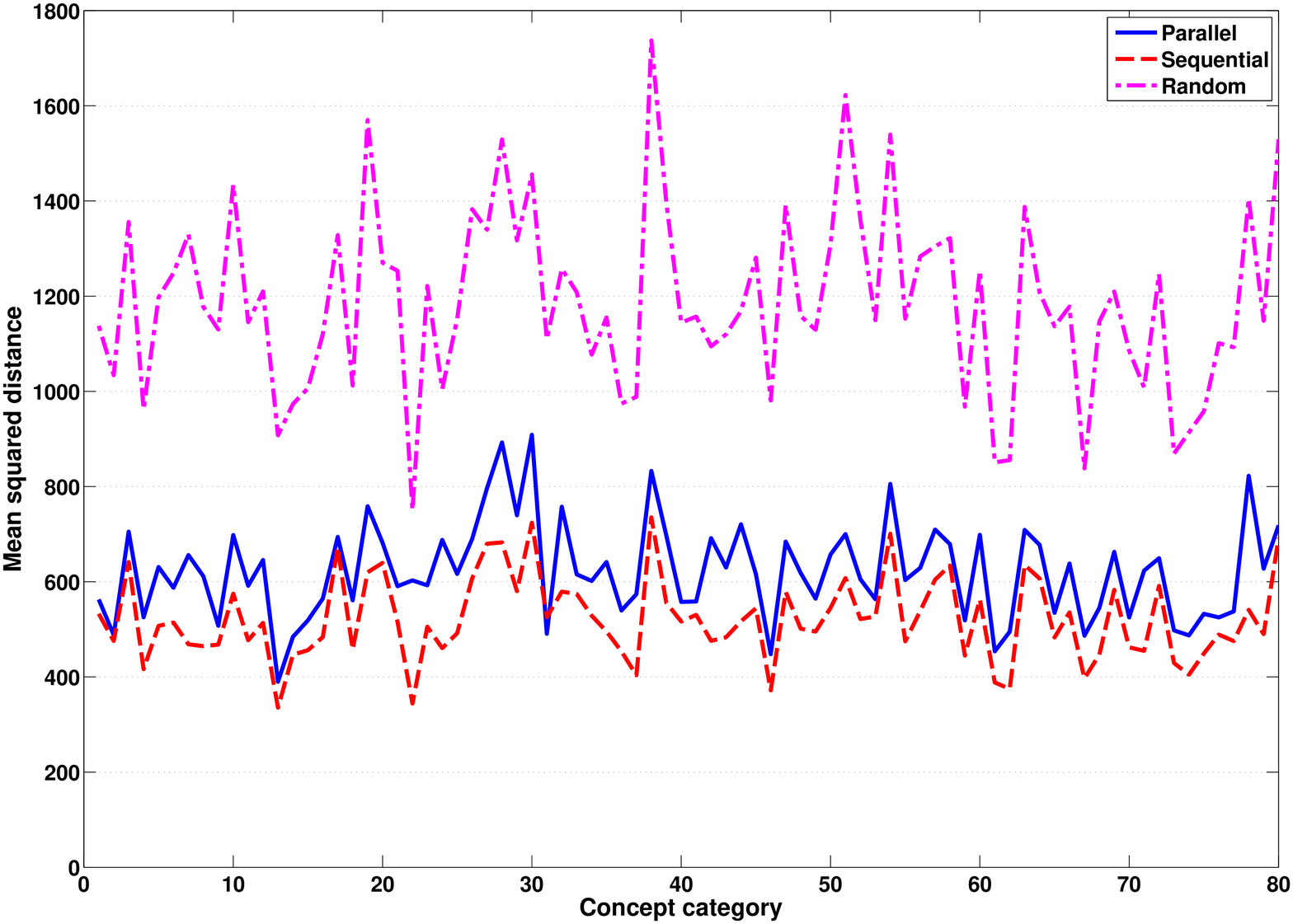}
	}
	\subfigure[]{
		\label{fig:cltcomp}
		\includegraphics[width=0.31\textwidth]{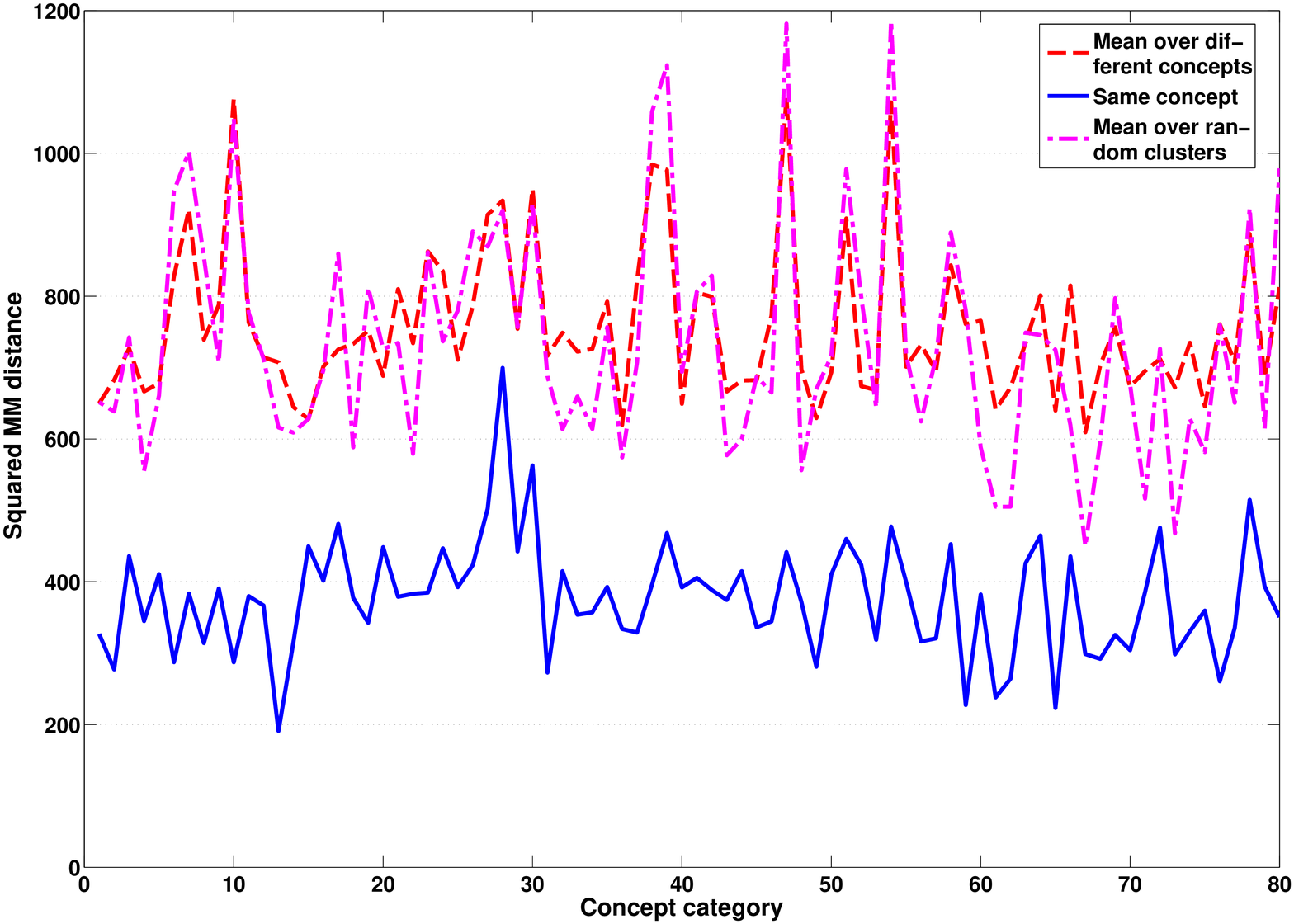}
	}
	\subfigure[]{
		\label{fig:labelcmp}
		\includegraphics[width=0.31\textwidth]{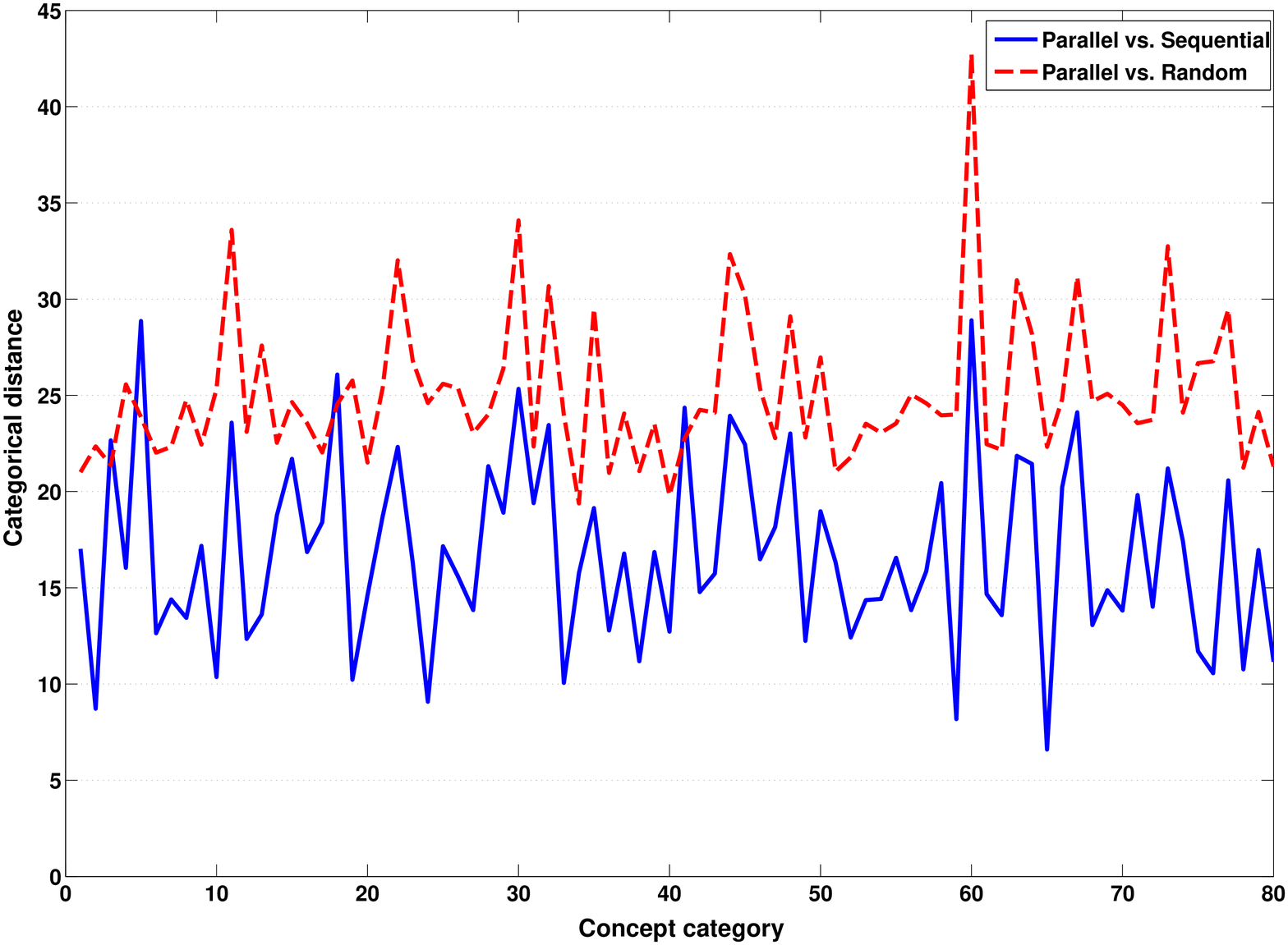}
	}
\vspace{-0.05in}
\caption{Comparison between parallel and sequential clustering.
(a) Mean squared distances from samples to corresponding
centroids. The clustering results acquired by parallel and
sequential D2-clustering algorithm on 80 sets are similar in
terms of mean squared distances to centroids.
(b) Comparison of the squared MM distances (\ref{equ:MM}) between
different sets of clustering centroids. Clustering centroids
derived from different approaches are compared.
(c) The categorical clustering distances~\cite{zhou2005new} between results
by different clustering approaches.}
\end{center}
\vspace{-0.1in}
\end{figure*}


\subsection{Comparison with Sequential Clustering} \label{sec:exp1}

In order to compare the clustering results of the parallel and
 sequential D2-clustering, we have to run the experiment on relatively
 small datasets so that the sequential D2-clustering can finish in a
 reasonable time.  In this experiment, 80 concepts of images, each
 including 100 images downloaded from Flickr, are clustered
 individually using both approaches.  We compare the clustering
 results obtained for every concept.

It is in general challenging to compare a pair of clustering results,
as shown by some sincere efforts devoted to this problem~\cite{zhou2005new}.  The difficulty comes from the need to match
clusters in one result with those in another.  There is usually no
obvious way of matching.  The matter is further complicated when the
numbers of clusters are different.  The parallel and sequential
clustering algorithms have different initializations and work flows.
Clearly we cannot match clusters based on a common starting point
given by initialization.  Moreover, the image signatures do not fall
into highly distinct groups, causing large variation in clustering
results.

We compare the clustering results by three measures.  

First, we assess the quality of clustering by the generally accepted
criterion of achieving small within-cluster dispersion, specifically,
the mean squared distance between a data entry and its corresponding \mbox{centroid}.

Second, we define a distance between two sets of centroids.  Let the
first set of centroids be $Z=\{z_1, ..., z_{k}\}$ and the second
set be $Z'=\{z'_1, ..., z'_{k'}\}$.  A centroid $z_i$ (or $z'_i$) is
associated with a percentage $p_i$ (or $p'_i$) computed as the
proportion of data entries assigned to the cluster of $z_i$ (or
$z'_i$).  We define a Mallows-type distance $\tilde{D}(Z,Z')$ based on
the element-wise distances $D(z_i, z'_j)$, where $D(z_i, z'_j)$ is the
Mallows distance between two centroids, which are both discrete
distributions.  We call $\tilde{D}(Z,Z')$ Mallows-type because it is
also the square root of a weighted sum of squared distances between
the elements in $Z$ and $Z'$.  The weights satisfy the same set of
constraints as those in the optimization problem for computing the
Mallows distance.
\vspace{-0.05in}
\begin{equation}
	\begin{aligned}
		&\tilde{D}^2(Z,Z')=\sum_{i=1}^{k}\sum_{j=1}^{k'}w_{i,j}D^2(z_i, z'_j)\;,\\
		&\text{subject to:}
		\sum_{j=1}^{k'}w_{i,j}=p_i\;,
		\sum_{i=1}^{k}w_{i,j}=p'_j\;,\\
		&w_{i,j}\geq 0, i=1,\ldots,k, j=1,\ldots,k'\;.
	\end{aligned}
	\label{equ:MM}
\vspace{-0.05in}
\end{equation}
We refer to $\tilde{D}(Z,Z')$ as the {\it MM distance} (Mallows-type weighted sum of
the squared Mallows distances).

Third, we use a measure for two ways of partitioning a dataset, which
is proposed in~\cite{zhou2005new}.  This measure only depends on the
grouping of data entries and is irrelevant with centroids.  We 
refer to it as the {\it categorical clustering distance}.  Let the first set
of clusters be $\mathcal{P}=\{P_1, P_2, ..., P_k\}$ and the second be
$\mathcal{P}'=\{P'_1, P'_2, ..., P'_{k'}\}$.  Then the element-wise
distance $D_p(P_i,P'_j)$ is the number of data entries that belong to
either $P_i$ or $P'_j$, but not the other.  The distance between the
partitions
$\tilde{D}_p(\mathcal{P},\mathcal{P}')=\sum_{i=1}^{k}\sum_{j=1}^{k'}
w_{i,j}D_p(P_i, P'_j)$ is again a Mallows-type weighted sum of
$D_p(P_i,P'_j)$, $i=1, \ldots, k$, $j=1,\ldots, k'$.  See~\cite{zhou2005new} for details.

For each concept category, the number of clusters is set to 10.  The
mean squared distance from an image signature to its closest centroid
is computed based on the clustering result obtained by the parallel or
sequential D2-clustering algorithm.  These mean squared distances are
plotted in Fig.~\ref{fig:sos}.  We also compute the mean squared
distance using randomly generated clusters for each concept.  The
random clustering is conducted 10 times and the average of the mean
squared distances is recorded as a baseline.  The random clusters,
unsurprisingly, have the largest mean squared distances.  In
comparison, the parallel and sequential algorithms obtain close values
for the mean squared distances for most concepts.  Typically the mean
squared distance by the parallel clustering is slightly larger than
that by the sequential clustering.  This shows that the parallel
clustering algorithm compromises the tightness of clusters slightly
for speed.

Fig.~\ref{fig:cltcomp} compares the clustering results using the MM
distance between sets of centroids acquired by different methods.  For
each concept $i$, the parallel algorithm generates a set of centroids
$Z_i=\{z_1,\ldots,z_{k_i}\}$ and the sequential algorithm generates a
set of centroids $Z'_i=\{z'_i,\ldots,z'_{k'_i}\}$.  The solid curve in
Fig.~\ref{fig:cltcomp} plots $\tilde{D}^2(Z_i,Z'_i)$ for each concept
$i$.  To demonstrate that $\tilde{D}(Z_i,Z'_i)$ is relatively small,
we also compute $\tilde{D}(Z_i,Z'_j)$ for all pairs of concepts $i\neq
j$.  Let the average $d^2_i=\frac{1}{c-1}\sum_{j=1,j\neq
i}^c\tilde{D}^2(Z_i,Z'_j)$, where $c=80$ is the number of concepts.
We use $d^2_i$ as a baseline for comparison and plot it by the dashed
line in the figure.  For all the concepts, $d_i$ is substantially
larger than $\tilde{D}(Z_i,Z'_i)$, which indicates that the set of
centroids $Z_i$ derived from the parallel clustering is relatively
close to $Z'_i$ from the sequential clustering.  Another baseline for
comparison is formed using random partitions.  For each concept $i$,
we create 10 sets of random clusters, and compute the average over the
squared MM distances between $Z_i$ and every randomly generated
clustering.  Denote the average by $\tilde{d}^2_i$, shown by the dashed
dot line in the figure.  Again comparing with $\tilde{d}^2_i$, the MM
distances between $Z_i$ and $Z'_i$ are relatively small for all
concepts $i$.

Fig.~\ref{fig:labelcmp} plots the categorical clustering distance
defined in~\cite{zhou2005new} between the parallel and sequential
clustering results for every image concept.  Again, we compute the
categorical clustering distance between the result from the parallel
clustering and each of 10 random partitions.  The average distance is
shown by the dash line in the figure.  For most concepts, the
clustering results from the two algorithms are closer than those from
the parallel algorithm and random partition.  However, for many
concepts, the categorical clustering distance indicates substantial
difference between the results of the parallel and sequential
algorithm.  This may be caused to a large extent by the lack of
distinct groups of images within a concept.

In summary, based on all the three measures, the clustering results
by the parallel and sequential algorithms are relatively close.

\subsection{Scalability Property}

In Section~\ref{sec:SpeedGain}, we derive
that the parallel D2-clustering runs in approximately linearithmic
time, while the sequential algorithm scales up poorly due to its
polynomial time complexity. In order to demonstrate this, we perform
clustering on sets of images with different sizes using both the
parallel and sequential D2-clustering with different conditions, and examine the time consumed.

\begin{figure}[t]
	\begin{center}
		\includegraphics[width=0.45\textwidth]{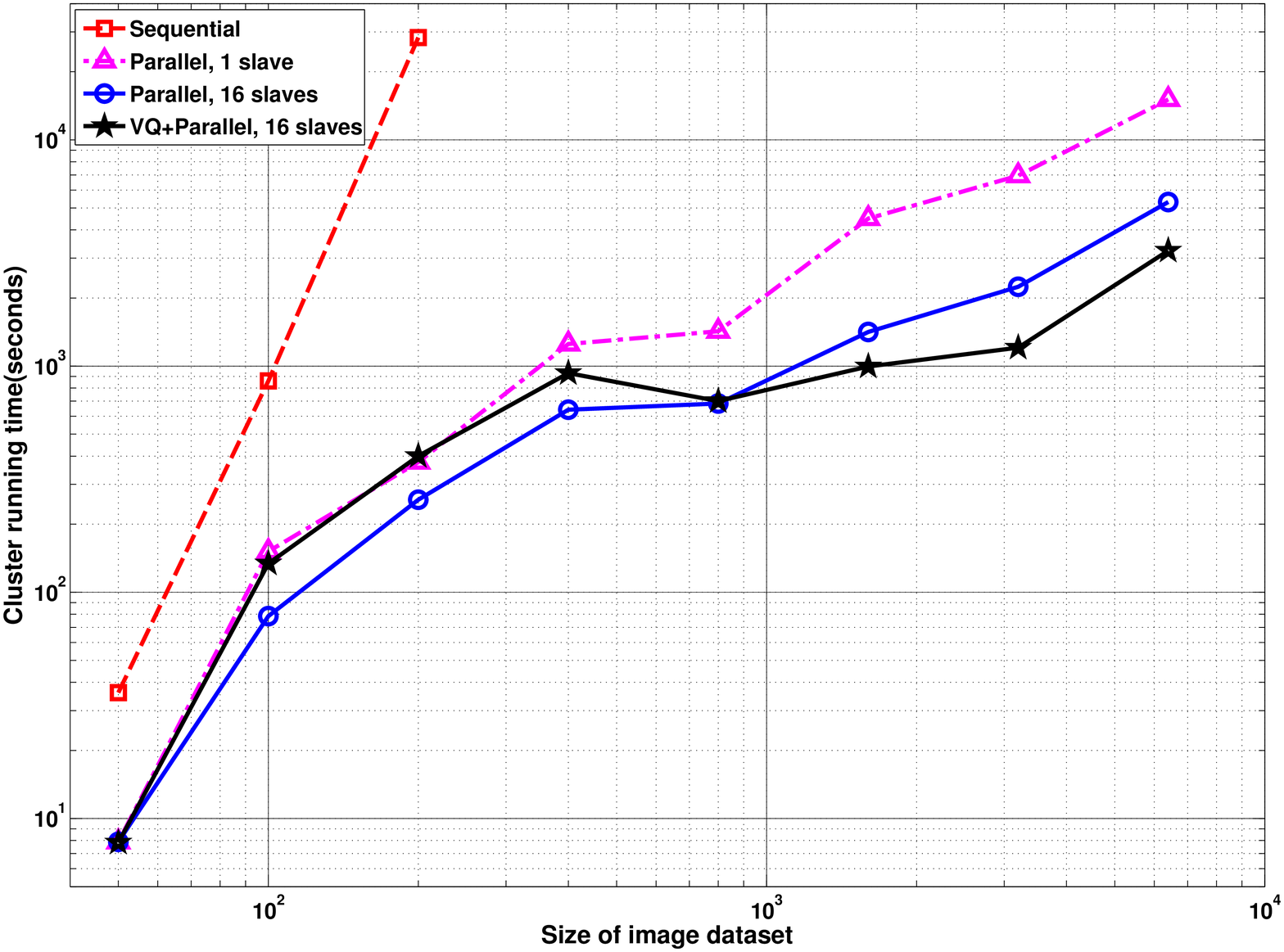}
	\end{center}
\vspace{-0.1in}
	\caption{Running time of clustering on different data size. Both axes are in logarithmic scales.
Both sequential and parallel D2-clustering algorithms are tested.
For the parallel algorithm there are three cases, depending on the number of slave
CPUs, and whether Vector Quantization (VQ) is adopted as the segmentation approach.
}
	\label{fig:time}
\vspace{-0.1in}
\end{figure}

Fig.~\ref{fig:time} shows the running time on sets of images in the
 concept ``mountain''.  In the plot, both axes are in logarithmic scales. 
All versions of clusterings are performed on datasets with sizes 50, 
100, 200, 400, 800, 1600, 3200, and 6400.
We test the parallel D2-clustering algorithm in three cases with identical parameters
$\tau=50$ and $e=5$. 
In the first case, there is only one slave CPU handling all the clustering
 requests sent by the master CPU (i.e. there are two CPUs employed in
 total).  All clustering requests are therefore processed sequentially
 by the only slave processor. 
In the second case, there are 16 slave CPUs (i.e. 17 CPUs in total).
In the third case, the conditions are the same to the second case, but the data segmentation
is implemented by a Vector Quantization (VQ) \cite{gersho1992vector} approach instead of
the approach in Section~\ref{sec:DataSegment}.
For comparison, the original sequential D2-clustering algorithm is also
tested on the same datasets. 

Fig.~\ref{fig:time} shows the parallel algorithm scales up approximately
 at linearithmic rate. On a log-log plot, this means the slope of the
 curve is close to 1. The number of slave CPUs for the parallel
 processing, as discussed in Section~\ref{sec:SpeedGain}, contributes a
linear difference (the second term in (\ref{equ:timecomp})) to the run-time
 between the first and second parallel cases. 
 When $N$ is small, the linear difference may be relatively significant 
 and make the single CPU case much slower than the multiple CPU case.
 But the difference becomes less so when the dataset is large 
 because there is a dominant linearithmic term in the run-time expression.
 Nevertheless, in either case, the parallel algorithm is much faster
 than the sequential algorithm.  The slope of the curve on the log-log
 plot for the sequential algorithm is much larger than 1, indicating
 at least polynomial complexity with a high degree.  As a matter of
 fact, the parallel algorithm with 16 slave CPUs takes 88 minutes to
 complete clustering on the dataset containing 6400 images, in
 contrast to almost 8 hours consumed by the sequential algorithm
 running only on the dataset of size 200.  Because the time for the
 sequential D2-clustering grows dramatically with the data size, we
 can hardly test it on dataset larger than 200.

For the last parallel case, we include VQ in the segmentation 
in order to further accelerate the algorithm.
Based on the analysis in Section \ref{sec:SpeedGain}, we know that data segmentation
takes a lot of time at each level of the algorithm. 
In the VQ approach, we quantify the bag of weighted vectors to histograms and treat
them as vectors.
These histograms are then segmented by K-means algorithm in the segmentation step. 
The clustering within each segment is still D2-clustering on the original data. However
the time spent for segmentation is much shorter than the original approach in Section~\ref{sec:DataSegment}.
In Fig.~\ref{fig:time} the acceleration is reflected by the difference between the run-time curves of
parallel D2-clustering with and without VQ segmentation when $N$ is large.
When $N$ is small and the data segmentation is not dominant in the run-time,
VQ will usually not make the clustering faster. In fact due to bad segmentation of such a coarse
algorithm, the time consumed for D2-clustering within each segment might be longer.
That is the reason why the parallel D2-clustering with VQ segmentation is slower than 
its counterpart without VQ segmentation (both have same parameters and 16 slave CPUs employed) 
in Fig.~\ref{fig:time} when $N$ is smaller than 800.

Theoretically VQ can reduce the order of the clustering from linearithmic to linear (referred by (\ref{equ:timecomp})). However because the quantization step loses some information, 
the clustering result might be less accurate.
This can be reflected by the MM distance (defined in (\ref{equ:MM})) 
between the parallel D2-clustering with VQ segmentation and the sequential D2-clustering results on a
dataset containing 200 images,
which is 19.57 on average for five runs. Compared to 18.41 as the average MM distance between 
the original parallel D2-clustering and sequential D2-clustering results, the VQ approach makes
the parallel clustering result less similar to the sequential clustering result which is regarded as the standard.

No matter whether VQ is adopted, the experiment shows the acceleration of parallel D2-clustering
over the sequential algorithm, which verifies the run-time analysis in Section~\ref{sec:SpeedGain}.
Applying the algorithm, we can greatly increase the scale of discrete distribution clustering problems.

\subsection{Visualization of Image Clustering}

\begin{figure}[t]
	\begin{center}
		\includegraphics[width=0.45\textwidth]{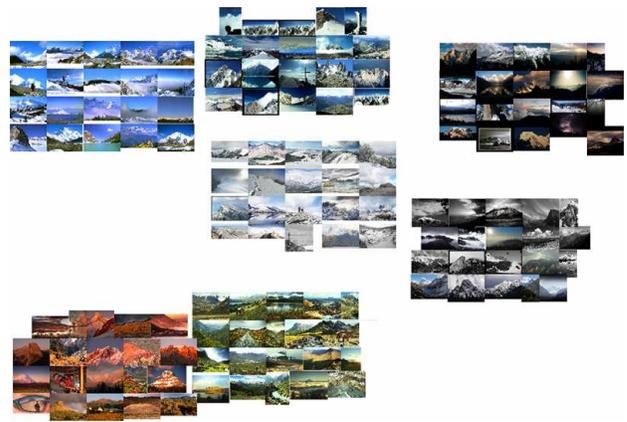}
	\end{center}
	\caption{Clustering result for 1488 images with concept ``mountain'' downloaded from Flickr. It takes 873 seconds to generate 13 clusters by using 30 slave CPU cores. Top 20 images from 7 largest clusters with size larger than 100 are displayed. The relative distance between two clusters represents the distance between their centroids.}
	\label{fig:mountain}
\vspace{-0.1in}
\end{figure}

Simulating the real image annotation application, 
the parallel D2-clustering algorithm is applied to a set of 1488 
images downloaded from Flickr ($\tau=50$, $e=5$).  These
images are retrieved by query keyword ``mountain''. For such a set, we
do not have ground truth for the clustering result, and the sequential
D2-clustering cannot be compared because of its unrealistic running
time.  We thus visualize directly the clustering result and attempt
for subjective assessment.

\begin{figure*}[t]
	\begin{center}
		\subfigure[Animals]{\includegraphics[width=3in]{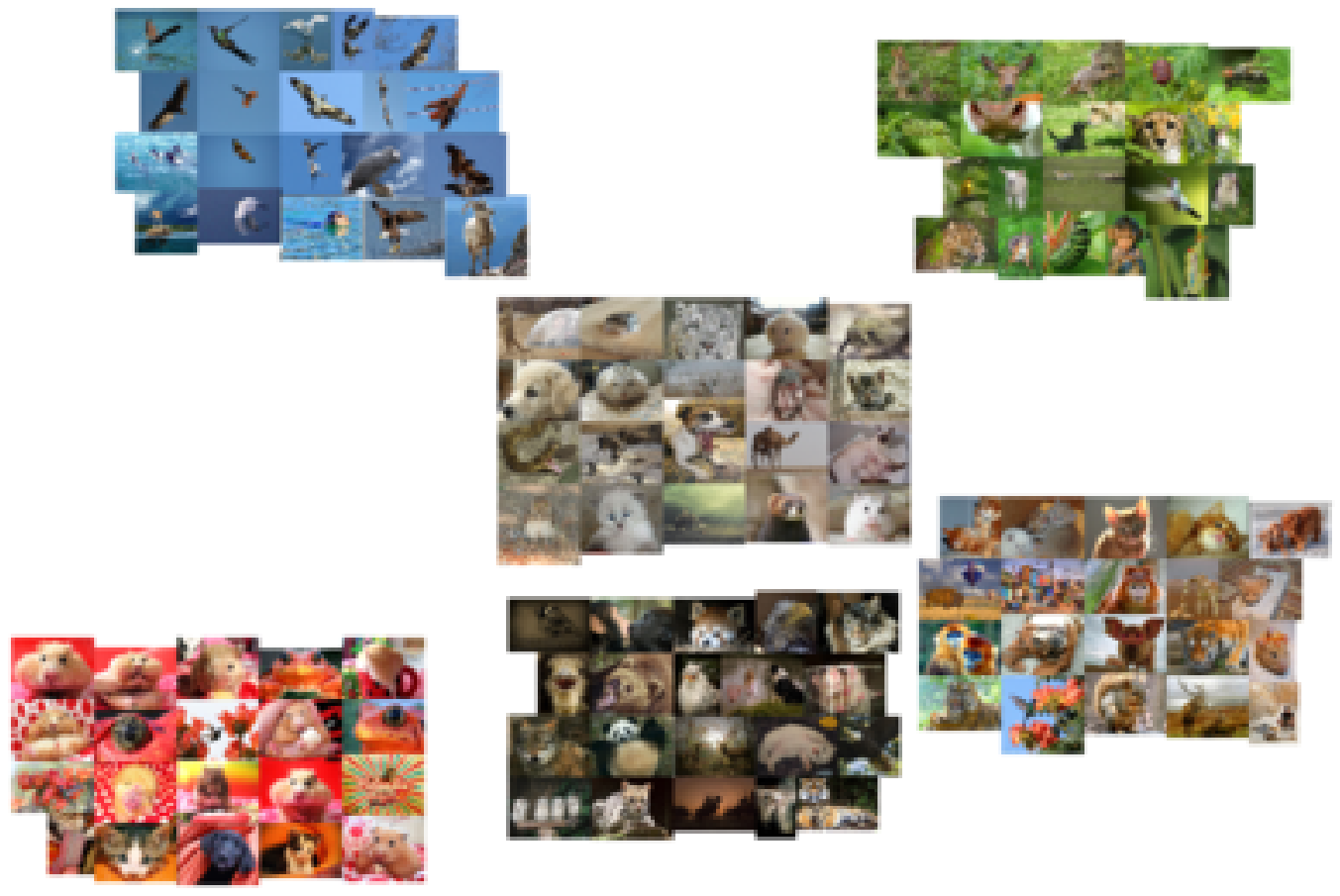}\hspace{0.3in}}
		\subfigure[China]{\includegraphics[width=3in]{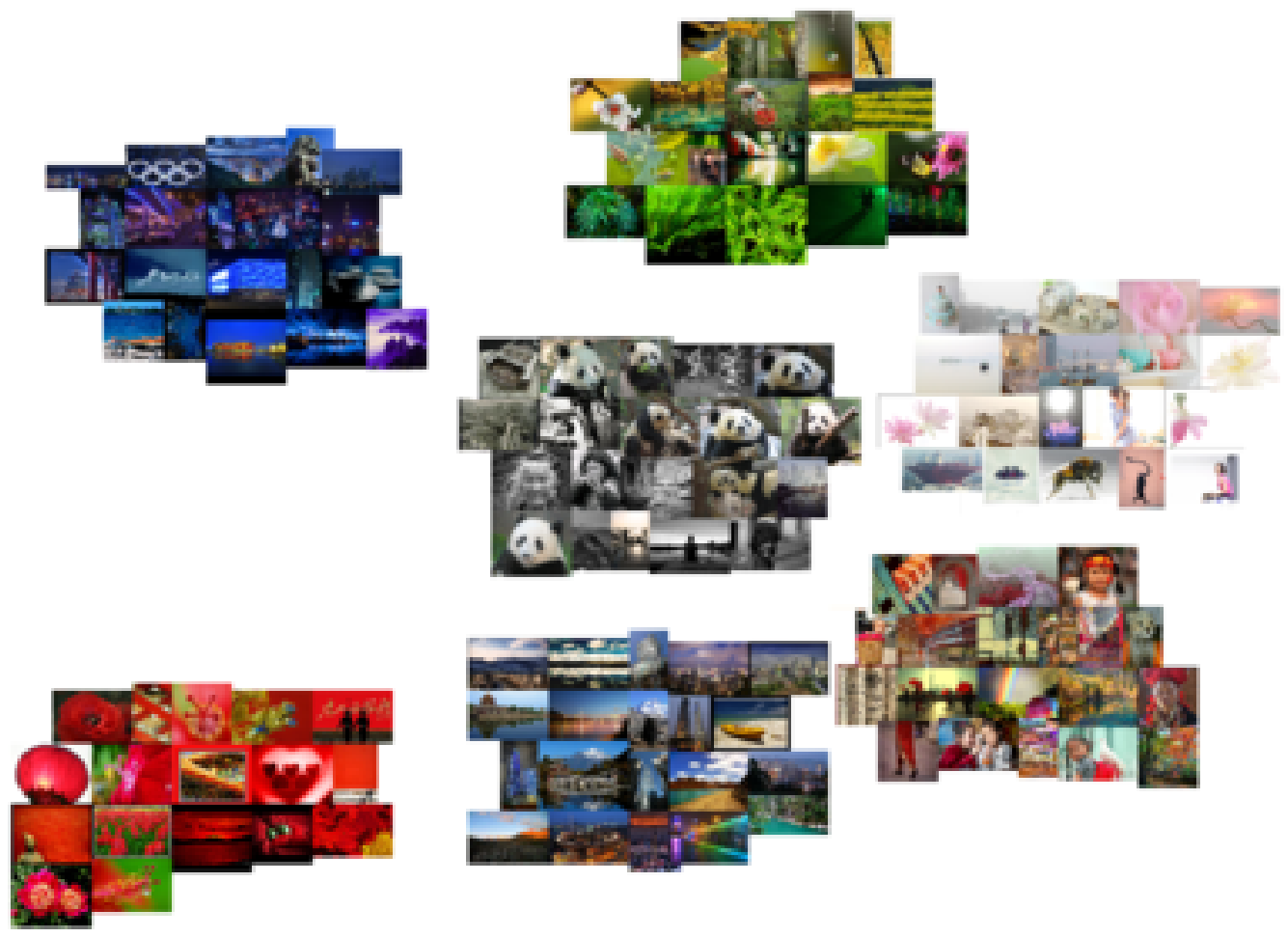}}
		\subfigure[Church]{\includegraphics[width=3in]{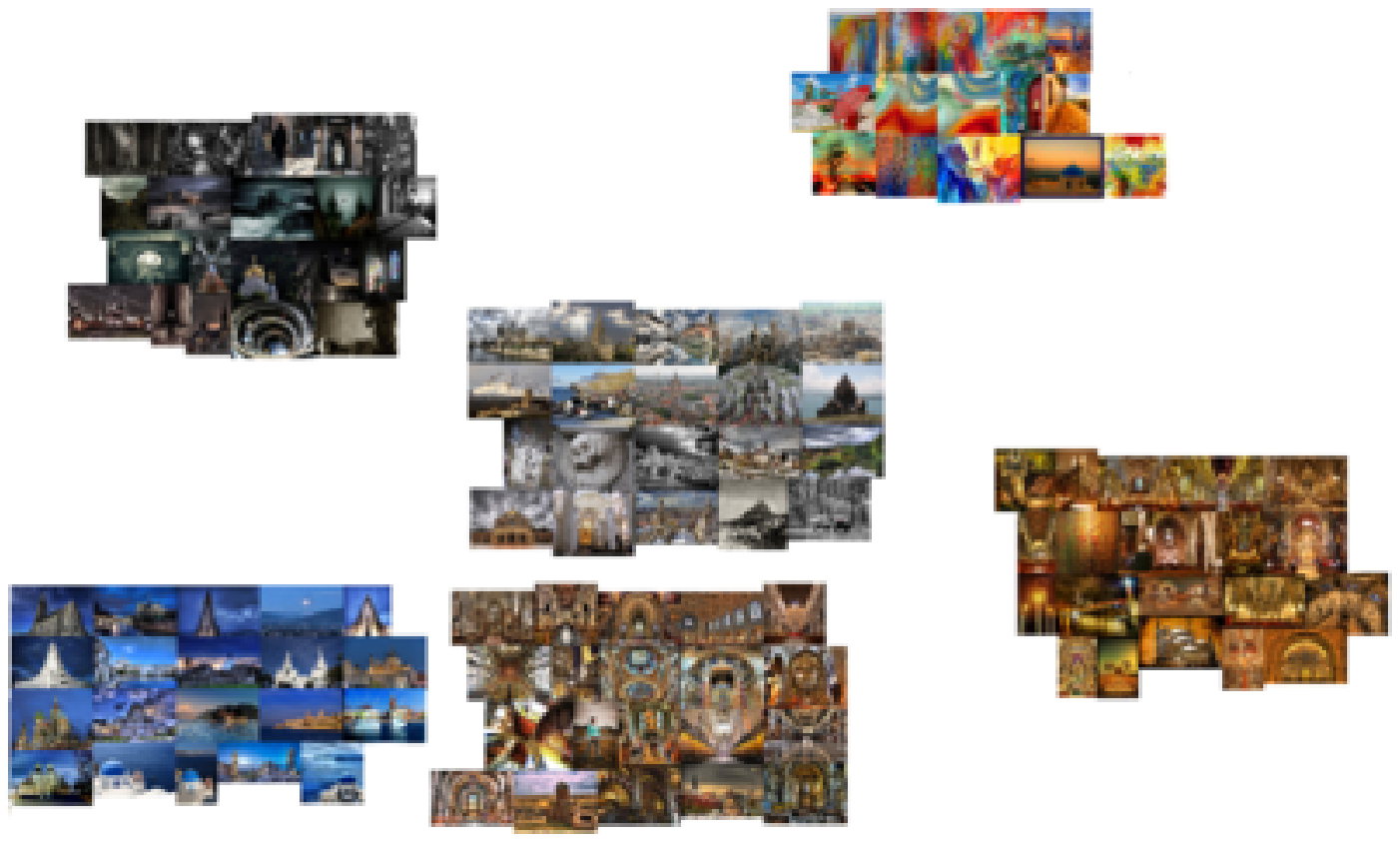}\hspace{0.3in}}
		\subfigure[Halloween]{\includegraphics[width=3in]{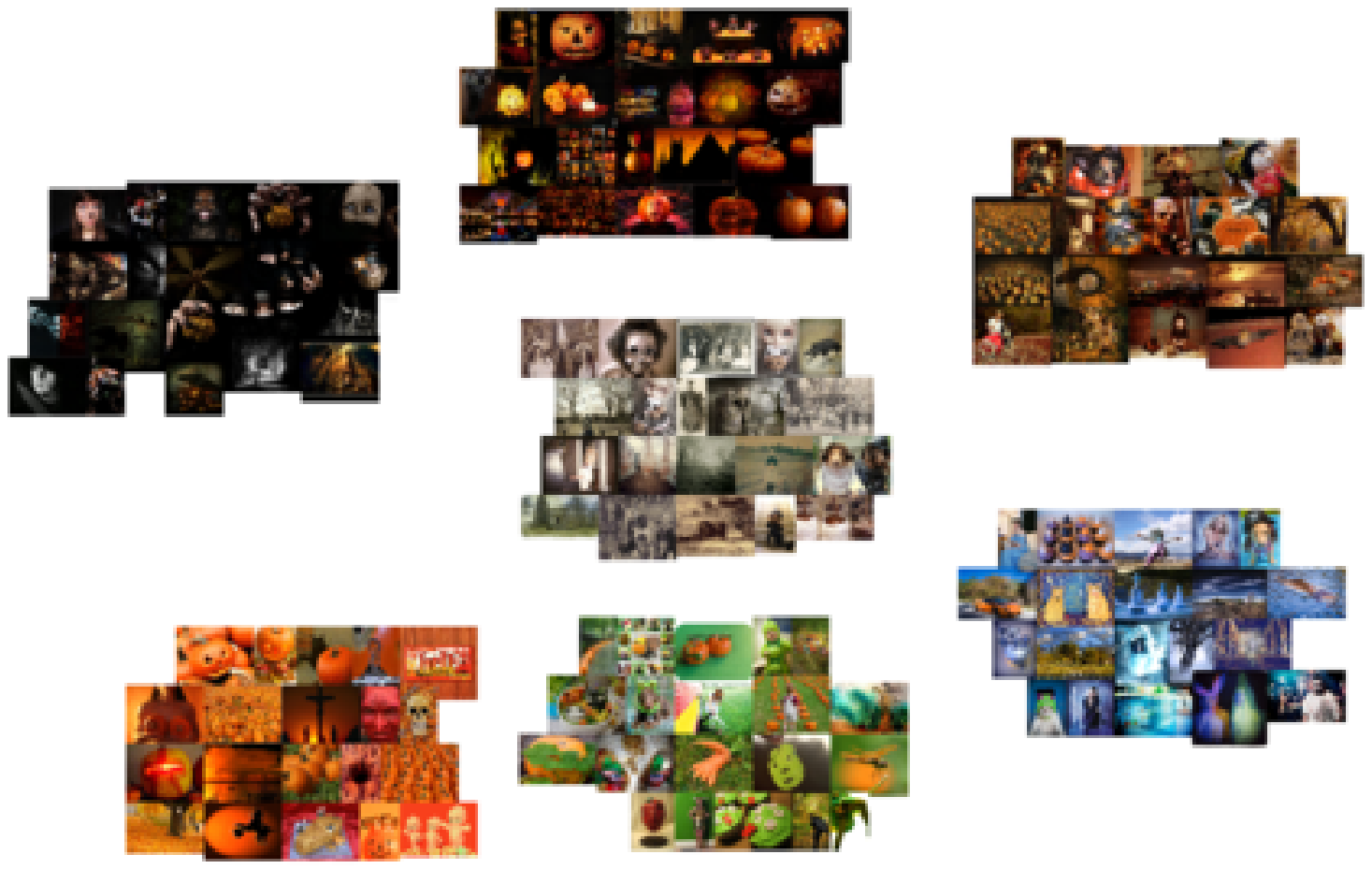}}
		
	\end{center}
\vspace{-0.1in}
	\caption{Visualized clustering results on four image concepts. Each concept contains 1,000 images from Flickr. They are clustered by the parallel D2-clustering algorithm and top ranked images in large clusters are displayed. Visually similar images are grouped into same clusters, though images in a same cluster do not always stand for similar semantics.}
	\label{fig:visual}
\vspace{-0.1in}
\end{figure*}

The 1488 images are clustered into 13 groups by the parallel
D2-clustering algorithm using 873 seconds (with 30 slave CPUs).  At the end, there are 7
clusters with sizes larger than 100, for each of which 20 images
closest to the corresponding cluster centroid are displayed in groups
in Fig.~\ref{fig:mountain}.  In the figure we try to manually arrange
similar clusters close, and dissimilar ones far away based on the
Mallows distances between the cluster centroids.
Fig.~\ref{fig:mountain} shows that images with similar color and
texture are put in the same clusters.  More clustering results for
images with different concepts are show in Fig.~\ref{fig:visual}.  For
all the datasets, visually similar images are grouped into the same
clusters.  The parallel algorithm has produced meaningful results.

\subsection{Video Clustering}
To demonstrate the application of parallel D2-clustering on video clustering, we download 450
videos from Youtube. They are all 5 to 20 minutes in length, and come from 6 different queries, which
are "news", "soccer", "lecture", "movie trailer", "weather forecast", and "tablet computer review". 
Each category contains 75 videos.
We compare the clustering result with the category label to check the correctness of the algorithm.

It should be emphasized that video classification is a complicated problem. Since our algorithm is an 
unsupervised clustering algorithm rather than a classification method, 
we cannot expect it to classify the videos with a high accuracy.
In addition, though not the major concern in this paper,
the visual features for segmenting and describing videos are crucial for the accuracy
of the algorithm. Here we only adopt some easy-to-implement simple features for demonstration.
Therefore the purpose of this experiment is not to pursue the best video classification, but to demonstrate
the reasonable results of the parallel D2-clustering algorithm on videos.
The videos' class labels, which serve as the
ground truth, are used as a reference to show whether similar 
videos can be clustered into a same group.

In this experiment, we set $\tau=30$, $e=5$, and request 15 slave CPUs ($M=15$) to cluster
the data. As mentioned above, we adopt three visual features, EDH, GBR, and GCD, to segment
and describe each video. We extract the features and form the video descriptors before
the clustering. Typically a video will be segmented into 10 to 30 sub-clips, depending
on its content and length. In addition, the feature vector itself is high in dimension (73 for EDH,
48 for GBR, and 225 for GCD). As a result, each video is represented by a large bag of weighted
vectors.

\begin{table}[h]
\caption{Video Clustering Result by Parallel D2-Clustering.}\label{tab:clt_video}
\vspace{-0.15in}
\begin{center}
\begin{tabular}{|c|c|c|c|c|c|c|}\hline
Cluster & C1 & C2 & C3 & C4 & C5 & C6 \\
(Size) & (81) &(36) & (111) & (40) & (72)& (75)  \\\hline\hline
Soccer& 43 & 5&1& 1  & 8& 10 \\\hline
Tablet& 19  & 23& 14 & 1  & 9&4 \\\hline
News & 9  & 3 &27& 6 & 17& 5 \\\hline
Weather& 4  & 0& 31 & 32 & 1 & 0 \\\hline
Lecture& 4  & 2 & 32& 0  & 17& 13 \\\hline
Trailer& 2  & 3 & 6& 0 & 20& 43 \\\hline
\end{tabular}
\end{center}
\vspace{-0.1in}
\end{table}

By applying the parallel D2-clustering algorithm, we can get a clustering result for this high dimensional
problem in 847 seconds. 6 major clusters (C1-C6) are generated and Table \ref{tab:clt_video} is the confusion table. 
We then try to use the clustering result for classification by assigning each cluster with the label corresponding
to its majority class. For these six clusters, 
 the unsupervised clustering achieves a classification accuracy of $47.5\%$.

\subsection{Protein Sequence Clustering}
In bioinformatics, composition-based methods for sequence clustering and classification 
(either DNA~\cite{kelley2010clustering, kislyuk2009unsupervised} or protein~\cite{garrow2005tmb})
use the frequencies of different compositions in each sequence as the signature.
Nucleotides and amino acids are basic components of DNA and protein sequence respectively
and these methods use a nucleotide or amino acid histogram to describe a sequence.
Because different nucleotide or amino acid pairs have different similarities
determined by their molecular structures and evolutionary relationships, cross-term
relations should be considered when we compute the distance between two such histograms.
As a result, the Mallows distance would be a proper metric in composition-based
approaches. However, due to the lack of effective clustering algorithms
under the Mallows distance, most existing clustering approaches either
ignore the relationships between the components in the histograms and treat them as variables in a vector,
or perform clustering on some feature vectors derived from the histograms.
In this case, D2-clustering will be especially appealing. Considering
the high dimensionality and large scale of biological data, parallel algorithm
is necessary. 

In this experiment, we perform parallel D2-clustering on 1,500 protein sequences
from Swiss-Prot database, which contains 519,348 proteins in total. The 1,500 proteins
are selected based on Prosite data, which provides class labeling for part of Swiss-Prot
database. We randomly choose three classes from Prosite data, and extract 500 protein
sequences from Swiss-Prot database for each class to build our experiment dataset.

Each protein sequence is transformed to a histogram of amino acid frequencies.
There is a slight modification in the computation of the Mallows
distance between two such histograms over the 20 amino acids.  In
Equation~(\ref{D2opt}), the squared Euclidean distance between two support
vectors is replaced by a pair-wise amino acid distance provided in the
PAM250 mutation matrix~\cite{pevsner2003bioinformatics}.
Given any two amino acids $A$ and $B$, we can get the probabilities of $A$ mutated to $B$ 
and $B$ mutated to $A$ from the matrix. The distance between $A$ and $B$ is defined as
\vspace{-0.05in}
\begin{equation}
	D_{\text{PAM250}}(A,B)=\log{(P(A|B))}+\log{(P(B|A))}\;.
	\label{equ:PAM250}
\vspace{-0.05in}
\end{equation}

\begin{figure}[t]
	\begin{center}
		\subfigure[Parallel D2-clustering]{\label{fig:seqclt}
\begin{minipage}[b]{\linewidth}
\centering
\includegraphics[width=0.3\textwidth, height=0.52in]{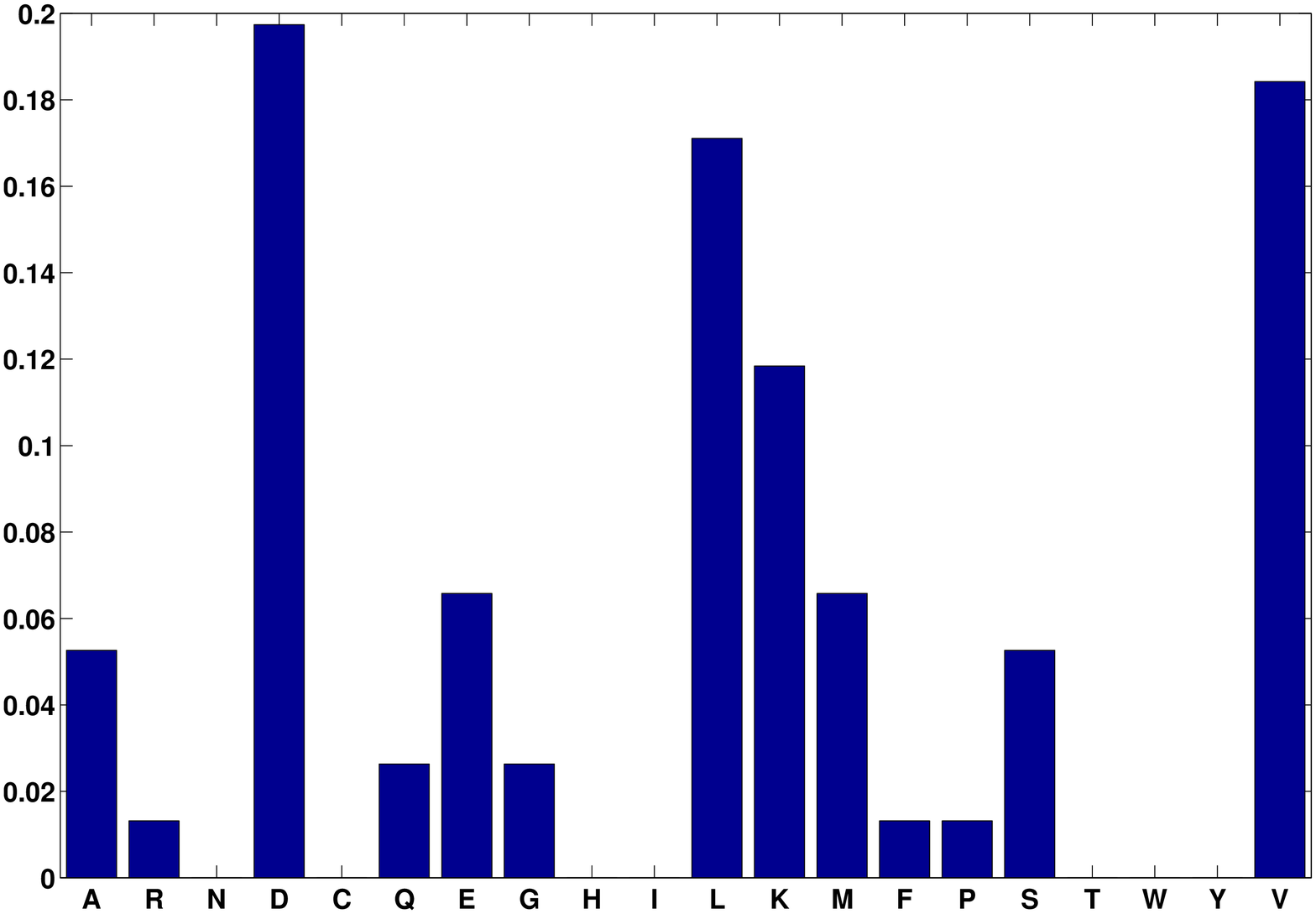}
\includegraphics[width=0.3\textwidth, height=0.52in]{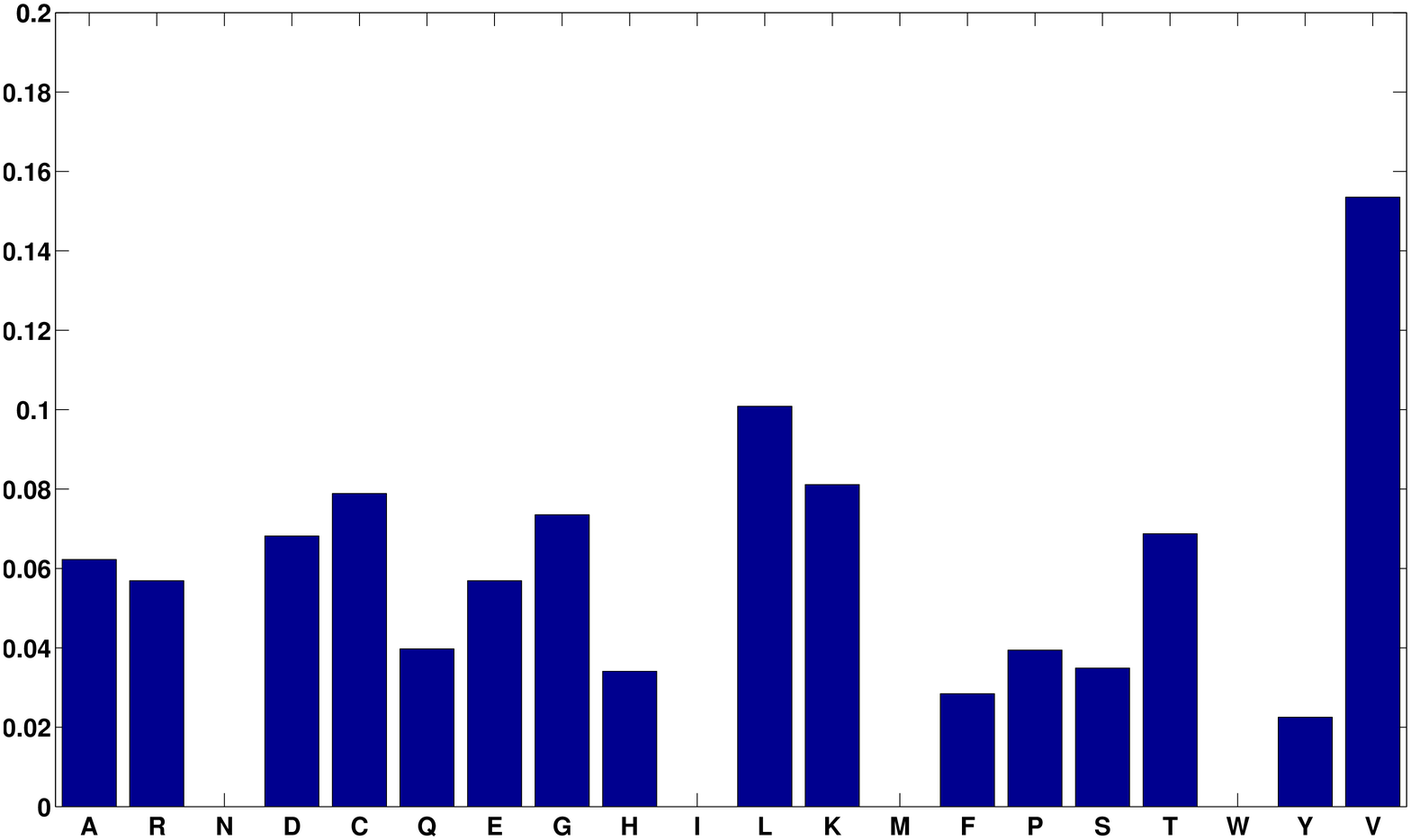}
\includegraphics[width=0.3\textwidth, height=0.52in]{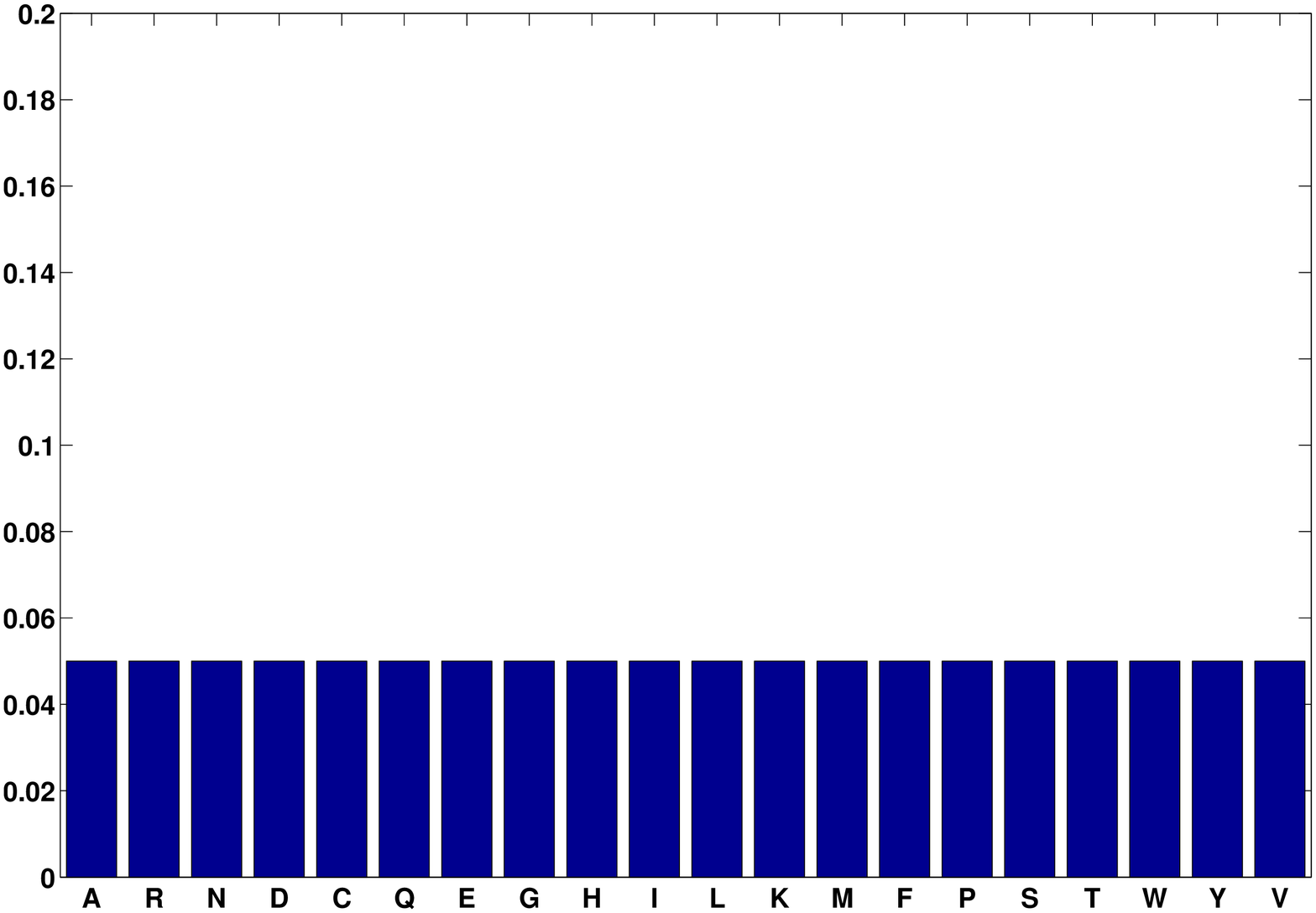}\\
{\footnotesize 50.80\%\hspace{0.2\textwidth}27.60\% \hspace{0.2\textwidth}21.47\%}
\end{minipage}}
		\subfigure[Parallel K-means]{\label{fig:seqclt_km}
\begin{minipage}[b]{\linewidth}
\centering
\includegraphics[width=0.3\textwidth, height=0.52in]{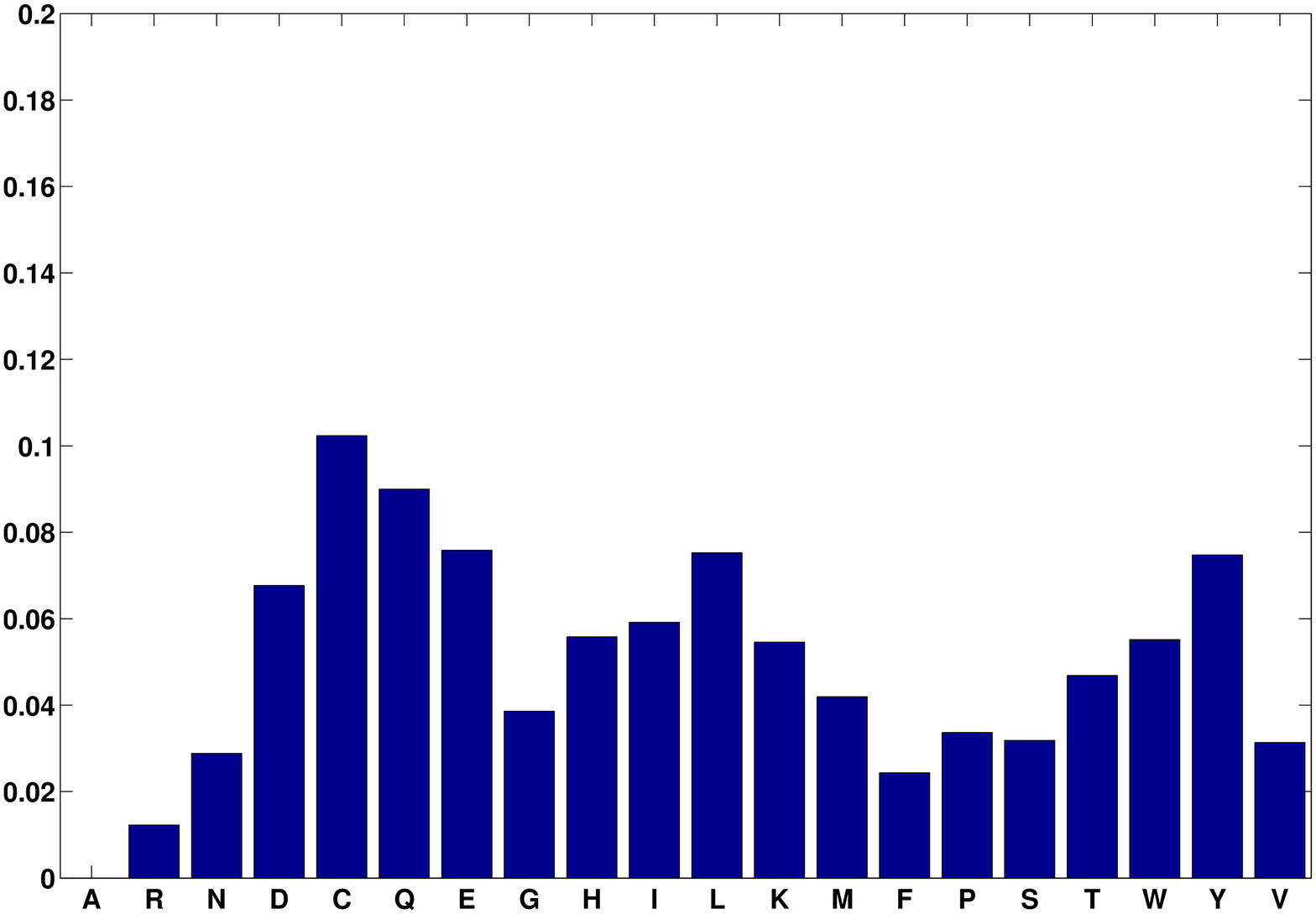}
\includegraphics[width=0.3\textwidth, height=0.52in]{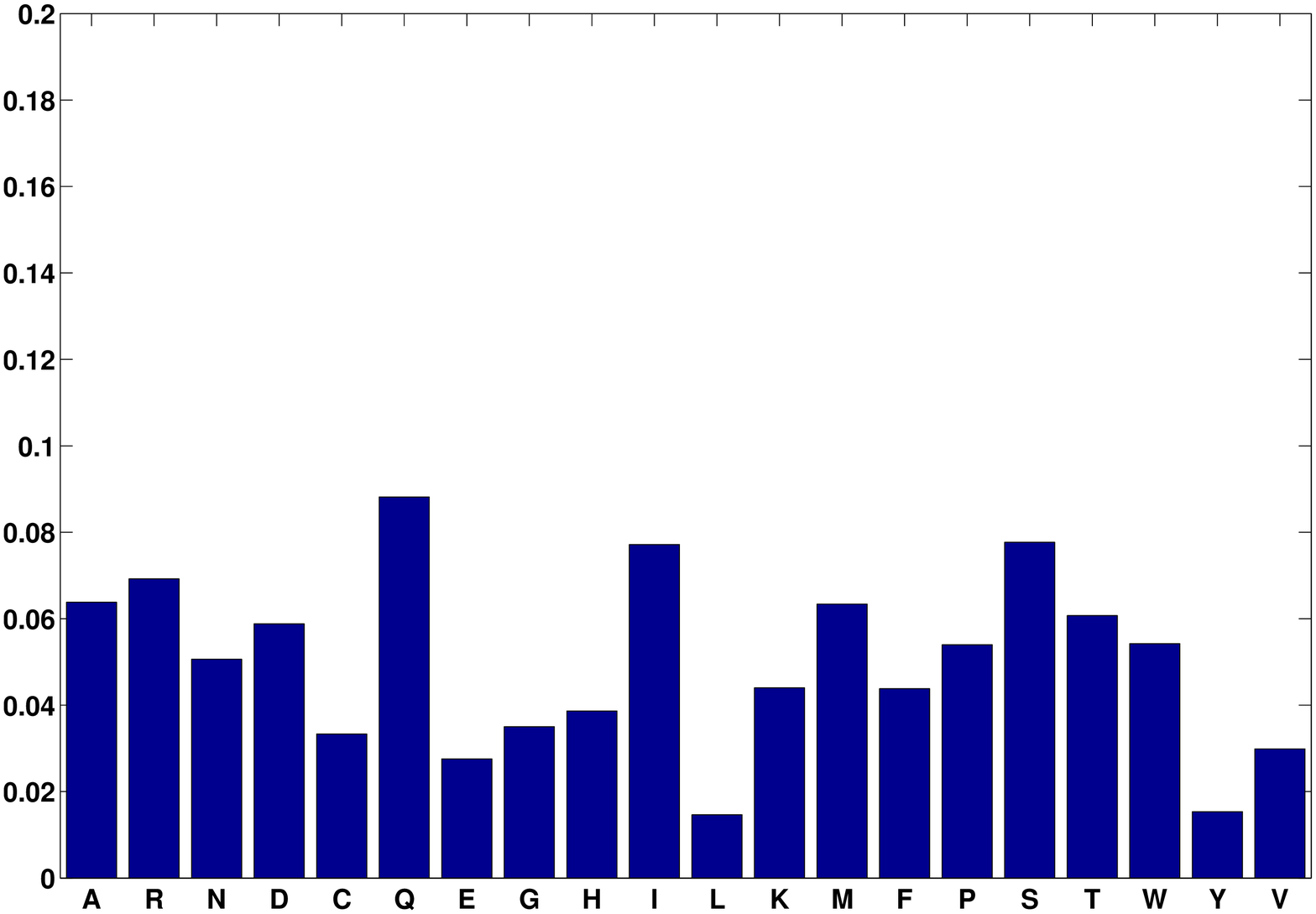}
\includegraphics[width=0.3\textwidth, height=0.52in]{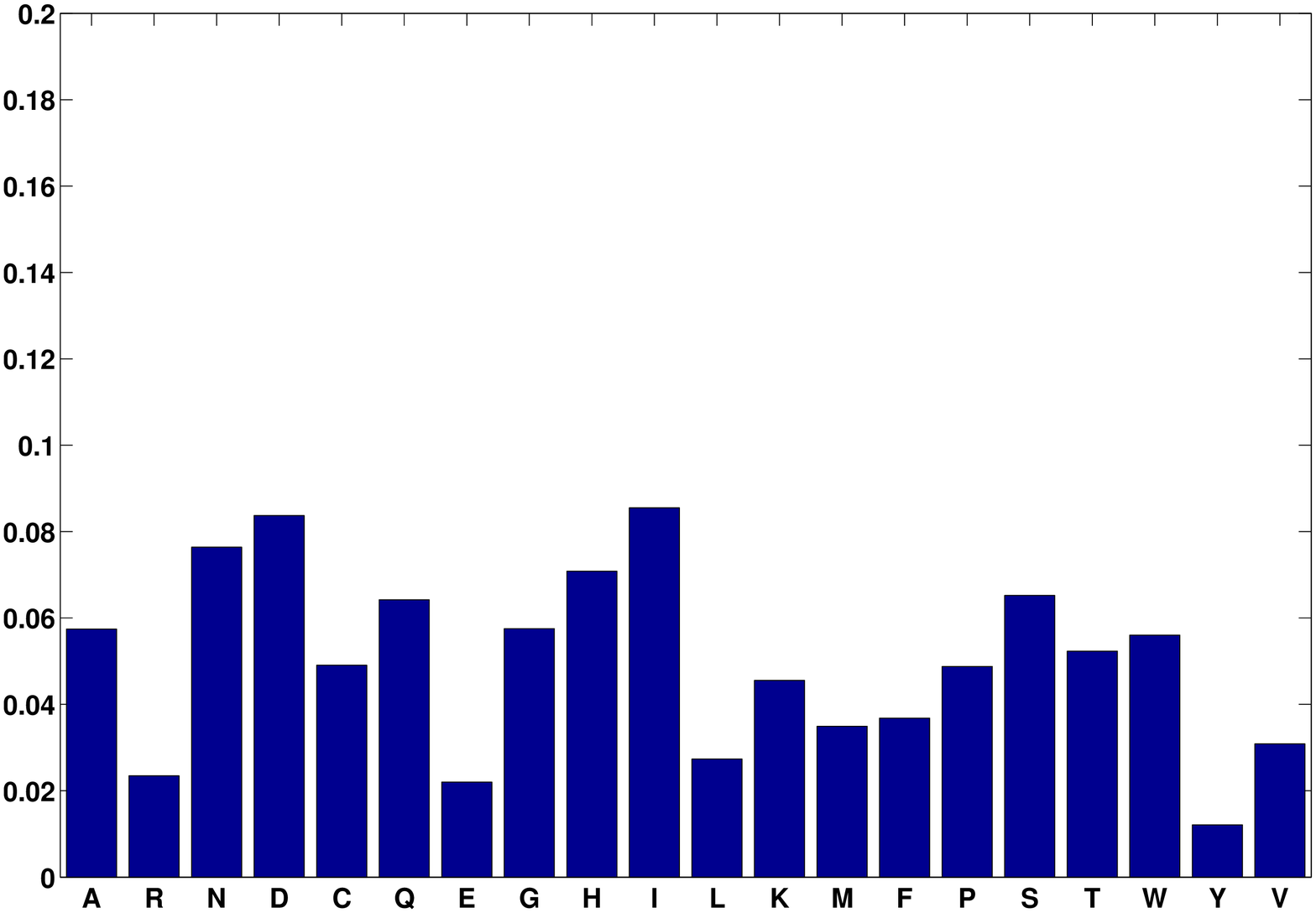}\\
{\footnotesize 51.00\%\hspace{0.2\textwidth}26.53\% \hspace{0.2\textwidth}22.47\%}
\end{minipage}}
		\subfigure[Hierarchical K-means]{\label{fig:seqclt_hk}
\begin{minipage}[b]{\linewidth}
\centering
\includegraphics[width=0.3\textwidth, height=0.52in]{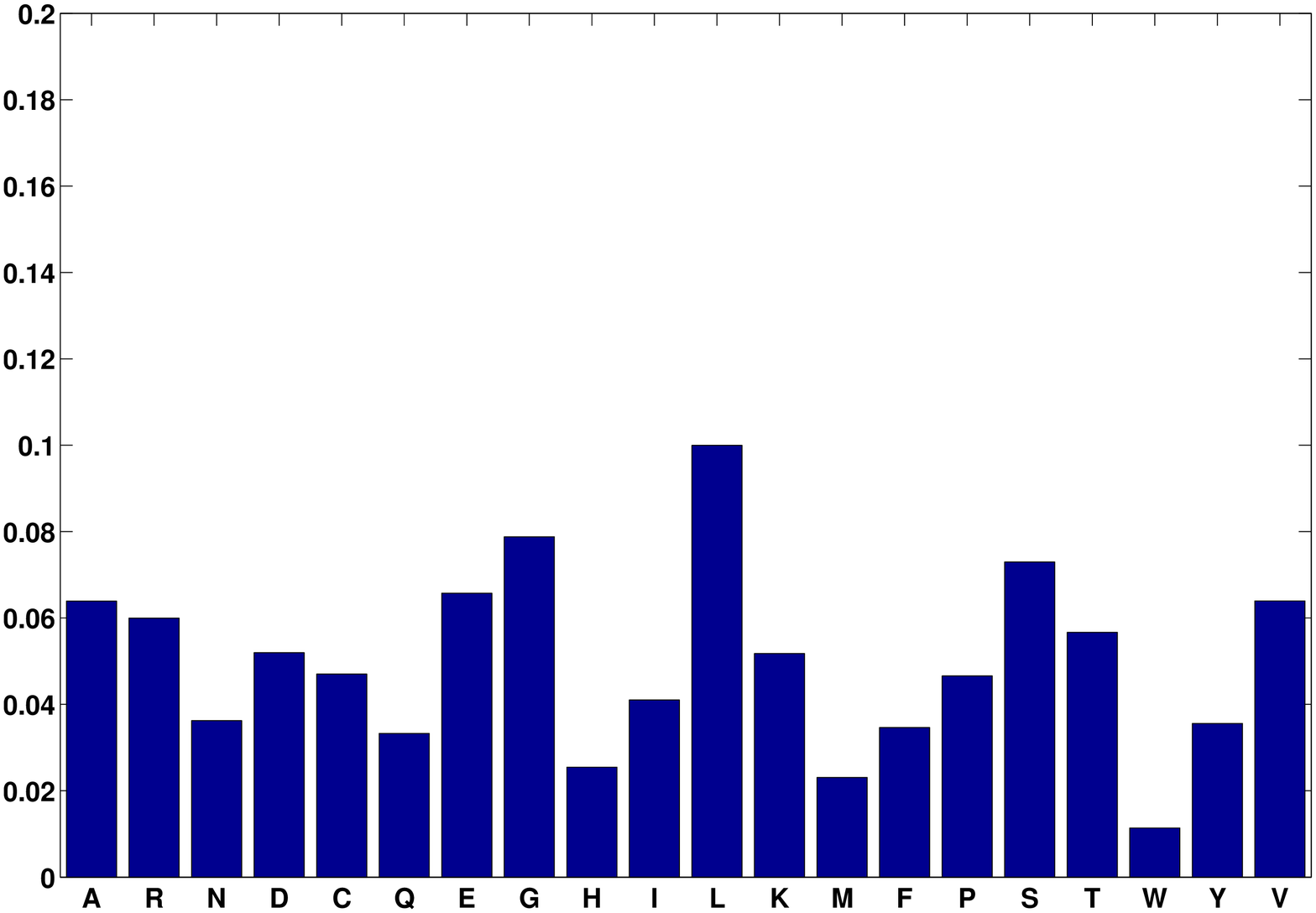}
\includegraphics[width=0.3\textwidth, height=0.52in]{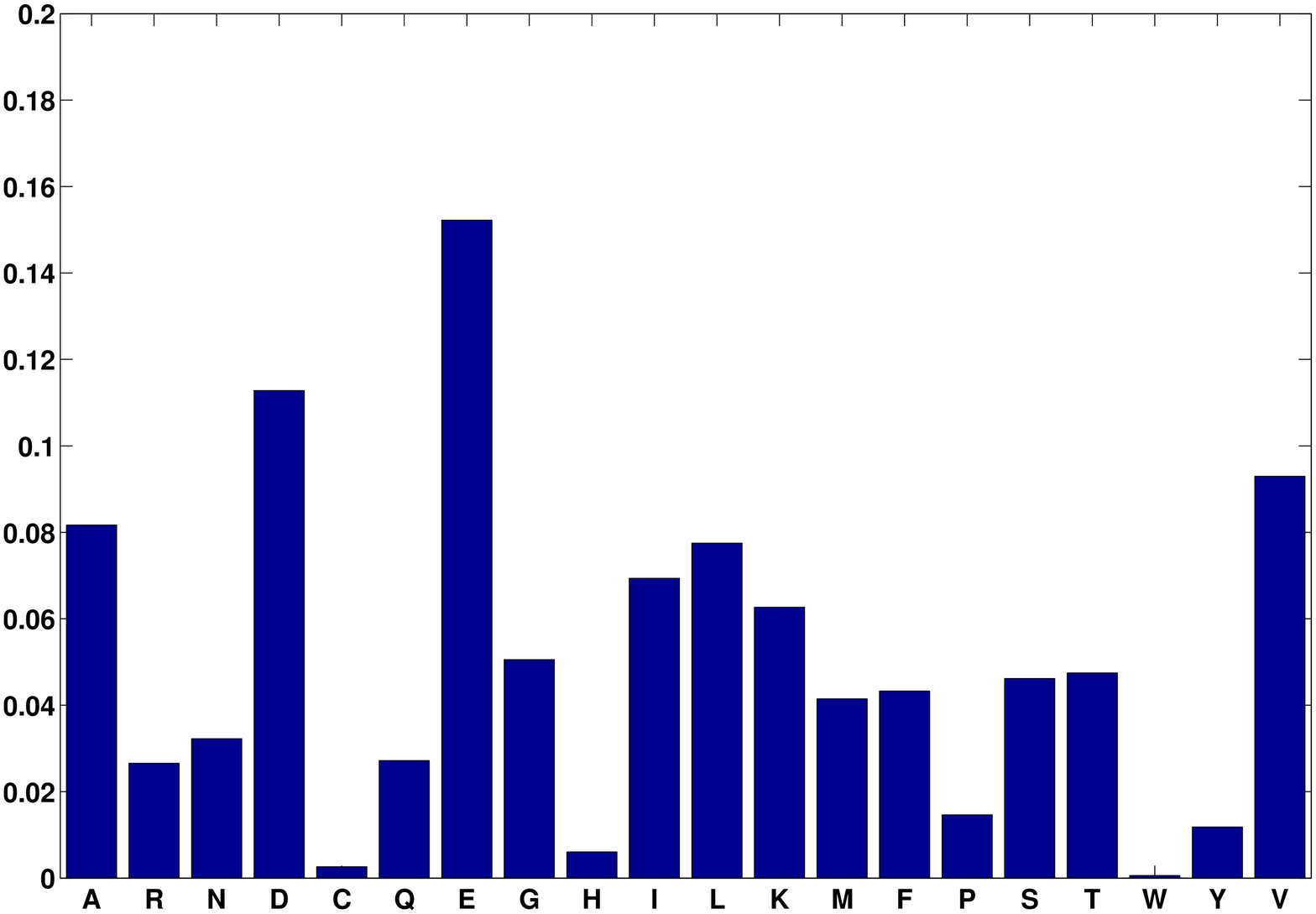}
\includegraphics[width=0.3\textwidth, height=0.52in]{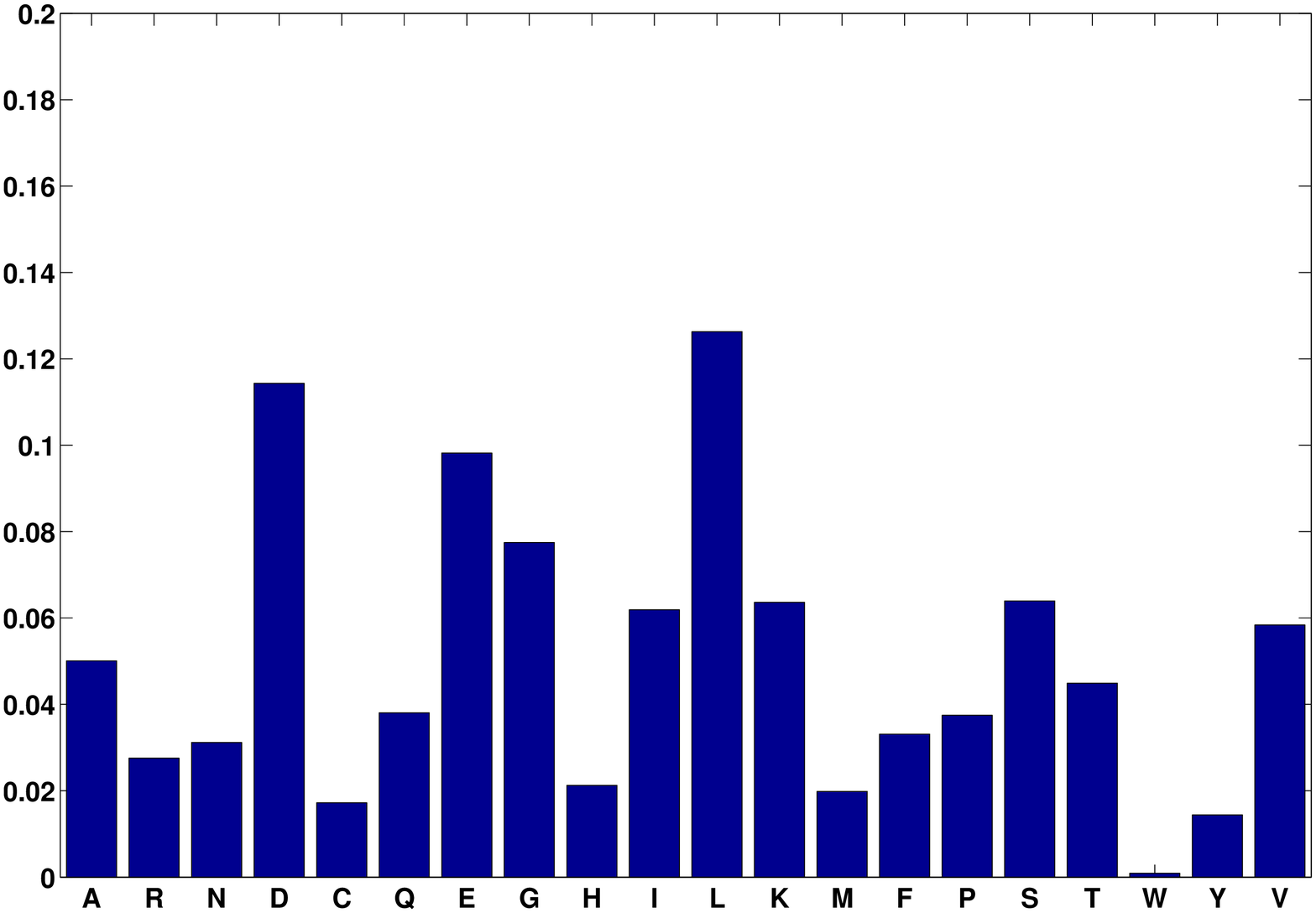}\\
{\footnotesize 49.00\%\hspace{0.2\textwidth}38.13\%\hspace{0.2\textwidth} 12.27\%}
\end{minipage}}
	\end{center}
\vspace{-0.1in}
	\caption{Cluster centroids computed by different approaches in the protein clustering experiment. Each plot, corresponding to one centroid, shows a histogram over the 20 amino acids. The number below each histogram
is the percentage of data points belonging to the corresponding cluster.}
\vspace{-0.1in}
\end{figure}

Because the support vectors for each descriptor  are the 20 types of amino acids and hence symbolic,
we will skip the step to update support vectors in the centroid update of D2-clustering (refer to Step~2 in Algorithm~1)
in the implementation of the parallel D2-clustering algorithm in this case. 
We set $\tau=30$, $e=5$, and request 7 slave processors to perform the parallel clustering ($M=7$).
We let the program generate 5 clusters and the clustering finishes in about 7 hours.
Considering the high dimensionality of the histograms and the scale of the dataset, this is
a reasonable time. Three out of the five clusters are major ones, which contain 
more than 99\% of the dataset. 
Fig.~\ref{fig:seqclt} describes the clustering centroids.

If we do not consider cross-term differences in the distance between two histograms
and use Euclidean distance as the metric, the clustering is reduced to K-means.
Therefore in this application, we are able to apply K-means on the dataset.
Running on the multi-CPU server, the K-means algorithm is also implemented
by parallel programs.
We parallelize K-means in two different ways. In the first one, we apply the approach
in ~\cite{kantabutra2000parallel}, which is to perform centroid update and label update
in parallel for different segments of data and then combine partial result linearly.
In this way the parallelization is equivalent to the original K-means on the whole dataset.
In the second way, we adopt the same hierarchical clustering structure, as in Fig. \ref{fig:parallelD2}, 
used in the parallel D2-clustering algorithm and only replace the Mallows distance by the Euclidean distance.
Thus within each data segment at any level of the hierarchy, the locally performed clustering is K-means.
However there will be some loss of accuracy since several data are grouped and treated as one object
at higher levels' clustering, and will never be separated thereafter. For short, we refer to the first
implementation as {\it parallel \mbox{K-means}} and the second one as {\it hierarchical K-means}.

Fig. \ref{fig:seqclt_km} and Fig. \ref{fig:seqclt_hk} show the clustering centroids generated by parallel
K-means and hierarchical K-means, running on the same dataset with identical parameters.
Apparently, none of parallel K-means' or hierarchical K-means' centroids reveals distinct patterns
compared with the centroids acquired by parallel D2-clustering. In order to prove the fact objectively,
we compute the Davies-Bouldin index (DBI)~\cite{davies1979cluster} for all the three clustering results
as the measurement of their tightness. DBI is defined as
\vspace{-0.05in}
\begin{equation}
	DB=\frac{1}{k}\sum_{j=1}^{k}{\max_{l\neq j}\left(\frac{\sigma_j+\sigma_l}{d(z_j,z_l)}\right)}\;,
	\label{equ:DBI}
\vspace{-0.05in}
\end{equation}
where $k$ is the number of clusters, $z_j$ is the centroid of cluster $j$,
$d(z_j, z_l)$ is the distance from $z_j$ to $z_l$,
and $\sigma_j$ is the average distance from $z_j$
to all the elements in cluster $j$.
DBI is the average ratio of intra-cluster dispersion to inter-cluster dispersion.
Lower DBI means a better clustering result with tighter clusters.

\begin{table}[h]
\caption{Davies-Bouldin index for different clustering results.}\label{tab:DBI}
\vspace{-0.15in}
\begin{center}
\begin{tabular}{|c||c|c|c|}\hline
Distance used in (\ref{equ:DBI})&\parbox{0.15\columnwidth}{Parallel D2}
&\parbox{0.15\columnwidth}{Parallel\\ K-means}
& \parbox{0.15\columnwidth}{Hierarchical K-means}\\\hline\hline
Squared Mallows & {\bf 1.55} & 2.99 &2.01\\\hline
Squared $L_2$ & {\bf 1.69} & 5.54 & 2.04
\\\hline
\end{tabular}
\end{center}
\vspace{-0.1in}
\end{table}

We compute DBI using both the squared Mallows distance and the
squared Euclidean distance as $d(\cdot)$ in~(\ref{equ:DBI})
for each clustering result. Table \ref{tab:DBI} shows the result. Parallel D2-clustering gains the lowest DBI in both cases.
This indicates good tightness of the clusters generated by parallel D2-clustering. In contrast, the two  implementations 
of \mbox{K-means} generate looser clusters than parallel D2-clustering. Though both can complete very fast, their clustering results are less representative.

\subsection{Evaluation with Synthetic Data}
Except for all the above experiments on real data, we also apply the
algorithm on a synthetic dataset. The synthetic data are bags of weighted
vectors. We carefully sample the vectors and their weights so that the data
gather around several centroids, and their distances to the corresponding
centroids follow the Gamma distribution, which means they follow a Gaussian
distribution in a hypothetical space ~\cite{li2007real}. 
By using this dataset,
we have some ideally distributed data with class labels.

We create a synthetic dataset containing $15$ clusters, each
composed of $100$ samples around the centroid. The clusters are well separated.
The upper bound for the number of clusters  in the algorithm is set 20.
It takes about 8 minutes to cluster the data (with 30 slave CPUs).  Then 17
clusters are generated. We try to match these 17 clusters to the ground truth
clusters: 9 of the 17 clusters are identical to their
corresponding true clusters; 5 clusters with sizes close to 100 differ
from the true clusters only by a small number of samples; 2 clusters
combine precisely into one true cluster; and at last a small cluster of size 8
may have formed from some outliers. It is evident that for this synthetic 
dataset, the parallel D2-clustering 
has produced a result close to the ground truth.

\section{Conclusion}\label{sec:conclude}
As a summary, a novel parallel D2-clustering algorithm with dynamic
hierarchical structure is proposed.  Such algorithm can efficiently
perform clustering operations on data that are represented as 
bags of weighted vectors. Due to the introduction of parallel computing, the
speed of the clustering is greatly enhanced, compared to the original
sequential D2-clustering algorithm. By deploying the parallel
algorithm on multiple CPUs, the time complexity of D2-clustering can
be improved from high-ordered polynomial to linearithmic time,
with minimal approximation introduced.
The parallel D2-clustering
algorithm can be embedded into various applications, including image
concept learning, video clustering, and sequence clustering problems.
In the future, we plan to further develop image concept learning and 
annotation applications based on this algorithm.

\ifCLASSOPTIONcompsoc
  \section*{Acknowledgments}
\else
  \section*{Acknowledgment}
\fi

This material is based upon work supported by the National Science
Foundation under Grant No. 1027854, 0347148, and 0936948. The 
computational infrastructure was provided by the Foundation through
Grant No. 0821527.  Part of the work of James Z. Wang and Jia Li is done while
working at the Foundation.

\ifCLASSOPTIONcaptionsoff
  \newpage
\fi

\bibliographystyle{IEEEtranS}
\bibliography{IEEEfull.bib,ref.bib}

\begin{IEEEbiographynophoto}{Yu Zhang}
received his B.S. degree in 2006 and M.S. degree in 2009, both in
Electrical Engineering, from Tsinghua University, Beijing.  He is
currently a Ph.D. candidate in the College of Information Sciences and
Technology at The Pennsylvania State University.
He has worked at Google Inc. at Mountain View as a software engineering intern
in both 2011 and 2012.
His research
interests are in large-scale machine learning, image retrieval,
computer vision, geoinformatics, and climate informatics.  He was awarded 
the academic excellence scholarship from Tsinghua University in 2004, and
the Graham Fellowship from The
Pennsylvania State University in 2009.
\end{IEEEbiographynophoto}

\begin{IEEEbiographynophoto}{James Z. Wang}
(S'96-M'00-SM'06)
received the B.S. degree in mathematics and computer science (summa cum laude) from the University
of Minnesota, Minneapolis, and the M.S. degree in mathematics, the M.S. degree in computer
science, and the Ph.D. degree in medical information sciences from Stanford University, Stanford,
CA. He has been with the College of Information Sciences and Technology of The Pennsylvania State University, University Park, since 2000, where he is currently a Professor and the Chair of the Faculty Council. From 2007 to 2008, he was a Visiting Professor of Robotics at Carnegie Mellon University, Pittsburgh, PA. He served as the lead
special issue guest editor for IEEE Transactions on Pattern Analysis and Machine Intelligence in 2008. In 2011 and 2012, he served as a Program Manager in the Office of the National Science Foundation Director. He has also held visiting positions with SRI International, IBM Almaden Research Center, NEC Computer and Communications Research Lab, and Chinese Academy of Sciences. His current research interests are automatic image tagging and retrieval, climate informatics, aesthetics and emotions, and computerized analysis of paintings. Dr. Wang was a recipient of a National Science Foundation Career Award and the endowed PNC Technologies Career Development Professorship.
\end{IEEEbiographynophoto}
\begin{IEEEbiographynophoto}{Jia Li}(S'94-M'99-SM'04)
is a Professor of Statistics and by courtesy appointment in
Computer Science and Engineering at The Pennsylvania State University,
University Park.  She received the M.S. degree in Electrical
Engineering, the M.S. degree in Statistics, and the Ph.D. degree in
Electrical Engineering, all from Stanford University. 
She has been serving as a Program Director in the Division of Mathematical Sciences
at the National Science Foundation since 2011.
She worked as a Visiting Scientist at Google Labs in
Pittsburgh from 2007 to 2008, a
Research Associate in the Computer Science Department at Stanford
University in 1999, and a researcher at the Xerox Palo Alto Research
Center from 1999 to 2000.  Her research interests include statistical
modeling and learning, data mining, signal/image
processing, and image annotation and retrieval.
\end{IEEEbiographynophoto}

\end{document}